\documentclass[journal]{IEEEtran}
\ifCLASSINFOpdf
\else
\fi

\usepackage{cite}
\usepackage{float}
\usepackage{graphicx}
\usepackage{multicol}
\usepackage{amssymb}
\usepackage{amsmath}
\usepackage{tikz}
\usepackage{algorithmic}
\usepackage{url}
\usepackage{listings}
\usepackage{pythonhighlight}
\usepackage{scalefnt}%
\newcommand\copyrighttext{%
  \footnotesize \textcopyright 2020 IEEE. Personal use of this material is permitted.
  Permission from IEEE must be obtained for all other uses, in any current or future
  media, including reprinting/republishing this material for advertising or promotional
  purposes, creating new collective works, for resale or redistribution to servers or
  lists, or reuse of any copyrighted component of this work in other works.}
\newcommand\copyrightnotice{%
\begin{tikzpicture}[remember picture,overlay]
\node[anchor=south,yshift=10pt] at (current page.south) {\fbox{\parbox{\dimexpr\textwidth-\fboxsep-\fboxrule\relax}{\copyrighttext}}};
\end{tikzpicture}%
}

\begin{document}
\copyrightnotice
%
% paper title
% Titles are generally capitalized except for words such as a, an, and, as,
% at, but, by, for, in, nor, of, on, or, the, to and up, which are usually
% not capitalized unless they are the first or last word of the title.
% Linebreaks \\ can be used within to get better formatting as desired.
% Do not put math or special symbols in the title.
\title{EMNIST Classification with Spiking Neural Network using SpykeFlow}
%
%
% author names and IEEE memberships
% note positions of commas and nonbreaking spaces ( ~ ) LaTeX will not break
% a structure at a ~ so this keeps an author's name from being broken across
% two lines.
% use \thanks{} to gain access to the first footnote area
% a separate \thanks must be used for each paragraph as LaTeX2e's \thanks
% was not built to handle multiple paragraphs
%

\author{Ruthvik~Vaila,~\IEEEmembership{Student~Member,~IEEE,}
        John~Chiasson,~\IEEEmembership{Fellow,~IEEE,}
        and~Vishal~Saxena,~\IEEEmembership{Member,~IEEE}% <-this % stops a space
\thanks{R. Vaila is a PhD candidate at Department
of Electrical and Computer Engineering, Boise State University, Boise,
ID, 83706 USA e-mail: ruthvikvaila@u.boisestate.edu}% <-this % stops a space
\thanks{J. Chiasson is with Boise State University, johnchiasson@boisestate.edu.}% <-this % stops a space
\thanks{V. Saxena is with University of Delaware, vsaxena@udel.edu.}}

\maketitle

% As a general rule, do not put math, special symbols or citations
% in the abstract or keywords.
\begin{abstract}
End user AI is trained on large server farms with data collected from the
users. With ever increasing demand for IOT devices, there is a need for deep
learning approaches that can be implemented (at the edge) in an energy
efficient manner. In this work we approach this using spiking neural networks.
The unsupervised learning technique of spike timing dependent plasticity
(STDP) and binary activations are used to extract features from spiking
input data. Gradient descent (backpropagation) is used only on the output
layer to perform the training for classification. The accuracies obtained for
the balanced \textsc{EMNIST} data set compare favorably with other approaches.
The effect of stochastic gradient descent (SGD) approximations on learning
capabilities of our network are also explored. We also introduce \textsc{SpykeFlow},
 a \textsc{Python} based software tool that we developed.

\end{abstract}

% Note that keywords are not normally used for peerreview papers.

% For peer review papers, you can put extra information on the cover
% page as needed:
% \ifCLASSOPTIONpeerreview
% \begin{center} \bfseries EDICS Category: 3-BBND \end{center}
% \fi
%
% For peerreview papers, this IEEEtran command inserts a page break and
% creates the second title. It will be ignored for other modes.
\IEEEpeerreviewmaketitle

\section{Introduction}
% The very first letter is a 2 line initial drop letter followed
% by the rest of the first word in caps.
% 
% form to use if the first word consists of a single letter:
% \IEEEPARstart{A}{demo} file is ....
% 
% form to use if you need the single drop letter followed by
% normal text (unknown if ever used by the IEEE):
% \IEEEPARstart{A}{}demo file is ....
% 
% Some journals put the first two words in caps:
% \IEEEPARstart{T}{his demo} file is ....
% 
% Here we have the typical use of a "T" for an initial drop letter
% and "HIS" in caps to complete the first word.
\IEEEPARstart{B}{iological} neurons communicate with each other by transmitting spikes which
are $70mV$ voltage pulses while artificial neural networks (ANNs) communicate
with each other using floating point computations. There are two popular
theories pertaining to how the information is encoded in the spiking input
image: rate coding and latency coding. Rate coding stipulates that the
information transfer from the input image to the next (hidden) layer is
embedded in the rate of spikes coming out of the input neurons. In this work
latency coding is used and it refers to the information in the image being
encoded in the relative spike times \cite{delorme2001} \cite{Kiselev}.
According to latency coding, earlier spikes (in time) carry more information
than later (in time) spikes \cite{delorme2001}. The synapses (weights) between
spiking neurons are modified according to spike timing dependent plasticity
(STDP), where the synapse is strengthened if an input neuron aides the output
neuron in spiking (arrives before the output neuron spikes) while the synapse
is weakened if an input neuron does not aide the output neurons in spiking
(arrives after the output neuron spikes) \cite{Markram} \cite{Masquelier2008}.
As STDP is an unsupervised learning rule, SNNs can be trained layer by layer.
The synapses (weights) in ANNs are modified using gradient descent
(backpropagation) to reduce the loss defined as an appropriate cost function
on the last (output) layer \cite{bottou2016}. More specifically, gradient
descent is used to update the weights of the network to an acceptable local
minimum of the cost on the output layer \cite{Lecun98}. In ANNs
input data is fed forward through the network and then the gradient of the
cost error is computed layer by layer going backwards to update the weights in
each layer. That is, the weights cannot be updated as one feed forwards the
input data giving rise to update locking problem \cite{frenkel2019}. This
makes backpropagation a global update rule unlike STDP which is a local update
rule \cite{Whittington}. Further, backpropagation uses the same weights for
the forward as well as the backward steps, which is referred to as the weight
transport problem \cite{GROSSBERG198723} \cite{Liao} \cite{Aidan}. Random
backpropagation (feedback alignment) was shown to mitigate this problem
\cite{Lillicrap}. A neuromorphic variant of the feedback alignment (random
backpropagation) was proposed in \cite{Neftci} and was shown to achieve an
accuracy of approximately $98\%$ on the \textsc{MNIST} dataset. Binary neural
networks are simply ANNs with binary weights and activations and have been shown to
achieve near state-of-the-art classification results with the \textsc{MNIST} \cite{mnist}
and \textsc{CIFAR-10} \cite{cifar10} datasets. Activations are binarized according to a
deterministic or stochastic binarization function, and as binary activations
are not differentiable, so called straight through estimators (STE) are used
\cite{courbariaux2016} for backpropagation. Panda et al. \cite{Panda} reported
a reduction in energy consumption by a factor of 25 for \textsc{CIFAR-10} and
reduction by a factor of 2 for the \textsc{ImageNet} dataset \cite{imagenet} by combining
existing techniques in deep learning with rate encoded spiking networks. Other
works like \cite{Anwani} \cite{Lee} \cite{Tavanaei2018} \cite{thiele2019}
approximate backpropagation with rate coding and have achieved approximately
$98\%$ accuracy on the \textsc{MNIST} dataset. Apart from training spiking
networks directly either with supervised or unsupervised methods, alternate
methods that convert an existing ANNs to SNNs using transfer learning was
introduced by \cite{Lungu}. Masquelier et al. \cite{Masquelier2017} proposed
that the earliest spike(s) are sufficient for rapid object classification. The
authors in \cite{comsa2019} \cite{Gardner} \cite{kheradpisheh2019s4nn}
\cite{mostafa2018} proposed an algorithm to learn exact spike times of
temporal coding (latency coding) with gradient descent (backpropagation) on an
output cost function. In their approach, activations in ANNs are replaced with
spike times and the loss is obtained by calculating the time difference
between the desired spike times and the actual spike times. In latency coding
minimizing a spike time is conceptually similar to maximizing the activation
of a target neuron.

The current literature on spiking networks indicate they give a lower accuracy
for classification \cite{panda2019} while standard ANNs employing SGD
(floating point computations) are energy inefficient due to the implementation
of the algorithms on high precision computers. Energy-efficiency (low power
consumption) and state-of-the-art classification accuracy are important goals.
In this paper, we approach this by combining STDP and approximate SGD with
binary activations to achieve near state-of-the-art classification accuracy on
\textsc{EMNIST} and \textsc{MNIST} datasets.
\section{Network Description\label{network_desc}}

\begin{figure}
\centering
\includegraphics[
height=1.75in,
width=3.5in
]%
{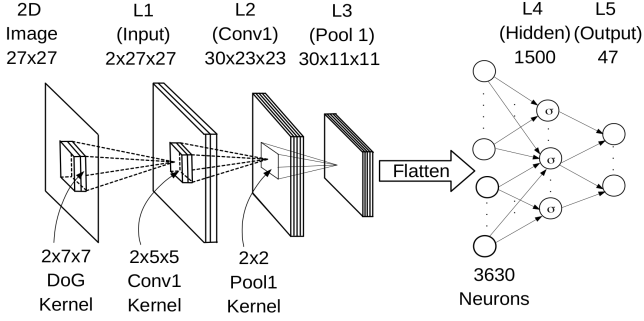}%
\caption{Layers $L1-L3$ are the feature extraction layers and layer $L3-L5$
are the feature classification layers. }%
\label{network1}%
\end{figure}

Our network is shown in Figure \ref{network1}. The extraction layers of this
network is similar to that of \cite{Kheradpisheh_2016}%
\cite{Kheradpisheh_2016b}.

%EndExpansion

\subsection{Input Encoding}

Following \cite{Kheradpisheh_2016} \cite{Kheradpisheh_2016b}, $K_{\sigma
_{1},\sigma_{2}}$ is a Difference of Gaussian (DoG) filter with $\sigma
_{1}=1,\sigma_{2}=2$ for the ON-center and $\sigma_{1}=2,\sigma_{2}=1$ for the
OFF-center, given by%

\begin{equation}
{\small
\begin{split}
 &K_{\sigma_{1},\sigma_{2}}(i,j)=\\
 &\begin{cases}
\dfrac{1}{2\pi\sigma_{1}^{2}}e^{-\dfrac{i^{2}+j^{2}}{2\sigma_{1}^{2}}}%
-\dfrac{1}{2\pi\sigma_{2}^{2}}e^{-\dfrac{i^{2}+j^{2}}{2\sigma_{2}^{2}}} &
\text{for }-3\leq i,j\leq3\\
& \\
0, \text{otherwise}%
\end{cases}
\end{split}
}
\end{equation}

\begin{figure}
\centering
\includegraphics[
height=1.25in,
width=3.0in
]%
{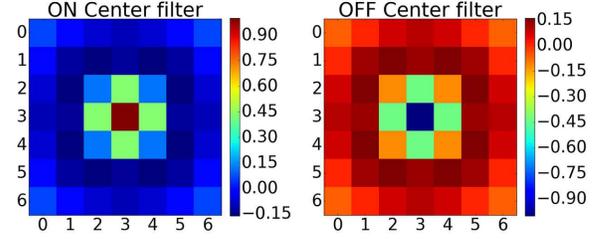}%
\caption{ON center filter has higher values at the center whereas OFF center
filter has lower values at the center. Colour code indicates the filter
values.}%
\label{onofffilters}%
\end{figure}

Plots of ON and OFF center filters are shown in Figure \ref{onofffilters}. The
input image is convolved with ON and OFF centered filters, resulting in two
\textquotedblleft images\textquotedblright\ which\ are then converted to an ON
and an OFF spiking image.%
\begin{equation}
{\small
\begin{split}
\Gamma_{\sigma_{1},\sigma_{2}}(u,v)=\sum_{j=-3}^{j=3}\sum_{i=-3}%
^{i=3}\mathbf{I}_{in}(u+i,v+j)K_{\sigma_{1},\sigma_{2}}(i,j) \\ 
\text{ \ for }0\leq u\leq26,0\leq v\leq26.
\end{split}
}
\end{equation}

At each location $(u,v)$ of the output image $\Gamma_{\sigma_{1},\sigma_{2}%
}(u,v)$ a unit spike $s_{(u,v)}$ is produced if and only if $\Gamma
_{\sigma_{1},\sigma_{2}}(u,v)$ exceeds a threshold i.e.,
\begin{equation}
\Gamma_{\sigma_{1},\sigma_{2}}(u,v)>\gamma_{DoG}%
\end{equation}
where $\gamma_{DoG}=50$ was chosen \cite{Kheradpisheh}. The spike times are
encoded relatively depending on magnitude of the membrane potentials and the
relation is given by%
\[
\tau_{(u,v)}=\frac{1}{\Gamma_{\sigma_{1},\sigma_{2}}(u,v)}\text{ in
milliseconds}.
\]
\bigskip%

\begin{figure}%
\centering
\includegraphics[
height=1.049in,
width=2.9378in
]%
{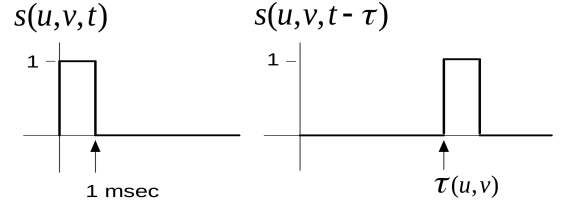}%
\caption{Left: Spike signal from the input image with no time delay. Right: Spike signal from the
input image with a delay of $\tau$ milliseconds.}%
\label{stdp_fig11b}%
\end{figure}

%EndExpansion
The spike signal $s_{(u,v)}(t)$ is latency (temporally) encoded
\cite{delorme2001} by delaying it by an amount inversely proportional to
$\Gamma_{\sigma_{1},\sigma_{2}}(u,v)$ as shown in Figure \ref{stdp_fig11b}.
That is, the greater the value of $\Gamma_{\sigma_{1},\sigma_{2}}(u,v)$, 
the sooner the neurons spikes and vice versa. Equivalently, the value of $\Gamma_{\sigma_{1},\sigma_{2}}(u,v)$ is
encoded in the value $\tau_{(u,v)}.$ Note that a neuron at location $(u,v)$
can generate at most one spike. Silicon retinas such as eDVS \cite{edvs} directly provide spiking images. The authors used such images in \cite{vaila2019}\cite{vaila2019a}.

%EndExpansion

\subsection{Convolution Layers and STDP}

We denote a spike at time $t$ emanating from the $(u,v)$ neuron of a spiking
image\ by $S_{L1}(t,k,u,v)$, where $k=0$ (ON center) or $k=1$ (OFF center) and
$(0,0)\leq(u,v)\leq(27,27)$. Layer $L2$ (Conv1) consists of 30 (feature) maps with
each map having its own convolution kernel given by%
\[
W_{C1}(w,k,i,j)\in%
\mathbb{R}
^{30\times2\times5\times5}\text{ \ for \ }w=0,1,2,...,29
\]
The instantaneous \textquotedblleft membrane potential\textquotedblright\ of
the $(u,v)$ neuron of a feature map $w$ ($w=0,1,2,...,29$) of layer $L2$ (Conv1) at
time $t$ is given by%
%\[
\begin{equation}
{\small
\begin{split}
&V_{L2}(t,w,u,v)=\\
&\sum_{t=0}^{\tau}\left(\sum_{k=0}^{1}\sum_{i=0}^{4}%
\sum_{i=0}^{4}S_{L1}(t,k,u+i,v+j)W_{C1}(w,k,i,j)\right) \\
&\text{\ \ for} \ 0\leq(u,v)\leq22
\end{split}
}
\end{equation}

If at time $t$ the membrane potential of a neuron in a feature map $w$ at
location $(u,v)$ crosses a set threshold value%
\[
V_{L2}(t,w,u,v)>\gamma_{L2}=15
\]
then the neuron at $(w,u,v)$ produces a spike at time $t$.

At any time $t,$ \emph{all} of the potentials $V_{L2}(t,w,u,v)$ for
$(0,0)\leq(u,v)\leq(22,22)$ and $w=0,1,2,...,29$ are computed in parallel.
Neurons in different locations within a map and in different maps may have
spiked. In particular, at the location $(u,v)$ there can be multiple spikes
(up to $30$) produced by $30$ different neuron belonging to $30$ different
maps. The desire is to have different maps learn different features so that
all the important features in the input image can be captured. To enforce this
condition, \emph{lateral inhibition} and \emph{STDP\ competition} are used
\cite{Kheradpisheh_2016}.

\subsubsection{Lateral Inhibition}

\begin{figure*}[!t]%
\centering
\includegraphics[
height=1.5in,
width=6.0in
]
{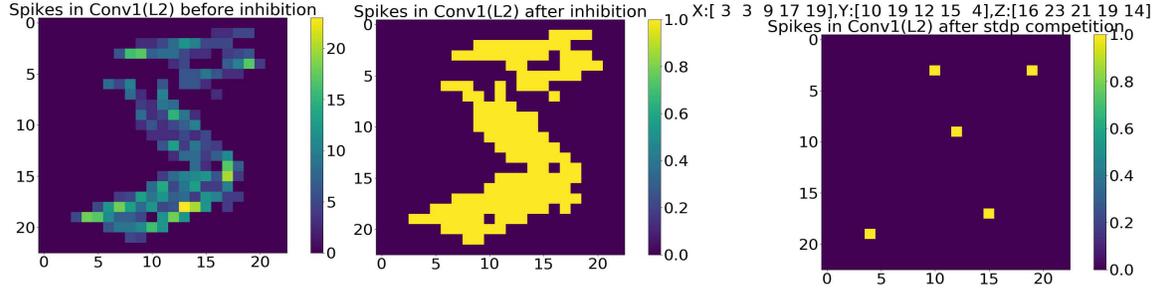}
\caption{Left: EMNIST digit "5" input. Accumulation of spikes from all 30 maps
and $12$ time steps in L2 \emph{without} lateral inhibition. Center:
Accumulation of spikes from all 30 maps and all $12$ time steps in L2
\emph{with} lateral inhibition. Right: Accumulation of spikes across all maps
and $12$ time steps with both lateral inhibition and STDP\ competition imposed
for a single image. X, Y denote the location of neuron in a map and Z denotes the map number. Note that these five winner spikes suppress all the other neurons that crossed the threshold.}
\label{l2ip2_inhbitions}%
\end{figure*}

To explain lateral inhibition suppose at the location $(u,v)$ there were
potentials $V_{L2}(t,w,u,v)$ in different maps ($w$ goes from $0$ to 29) at
time $t$ that exceeded the threshold $\gamma_{L2}.$ Then the neuron in a map with
the highest potential $V_{L2}(t,w,u,v)$ at $(u,v)$ inhibits the neurons in all
the other maps at the location $(u,v)$ from spiking for the current image
(even if the potentials in the other maps exceeded the threshold). Figure
\ref{l2ip2_inhbitions} (left) shows the accumulated spikes (from an
MNIST\ image of \textquotedblleft5\textquotedblright) for $12$ time steps from all 30 maps of
Layer $L2$ at each location $(u,v)$ \emph{without} lateral inhibition. For
example, at location (19,14) in Figure \ref{l2ip2_inhbitions} (left) the color
code is yellow indicating in excess of 20 spikes, i.e., more than 20 of the
maps produced a spike at that location.%

Figure \ref{l2ip2_inhbitions} (center) shows the accumulation of spikes from
all 30 maps for $12$ time steps, but now \emph{with} lateral inhibition imposed. Note that at each
location there is at most one spike indicated by the color code. Also, as
explained next, only a few of these spikes will actually result in the update
of any of the 30 kernels (weights) of layer L2.
\subsubsection{STDP\ Competition}

\begin{figure*}[!t]%
\centering
\includegraphics[
height=1.75in,
width=6.75in
]%
{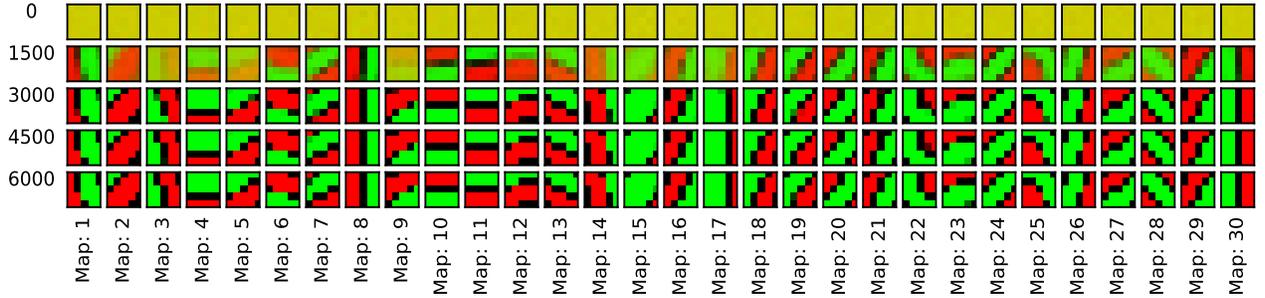}%
\caption{Evolution of learning in the first convolution layer. Red and Green indicate ON and OFF center synapses respectively.}%
\label{l1_filter_evolution}%
\end{figure*}

After lateral inhibition we consider each of the maps in layer L2 that had one
or more neurons with their potential $V$ exceeding $\gamma.$ Let these maps be
$w_{k1},w_{k2},...,w_{km}$ where\footnote{The other maps did not have any
neurons whose membrane potential crossed the threshold and therefore cannot
spike.} $0\leq k_{1}<k_{2}<\cdots<k_{m}\leq29$. Then in each map $w_{ki}$ we
locate the neuron in that map that has the maximum potential value. Let
\begin{equation}
(u_{k1},v_{k1}),(u_{k2},v_{k2}),...,(u_{km},v_{km}) \label{stdp1}%
\end{equation}
be the location of these maximum potential neurons in each map. Then neuron
$(u_{ki},v_{ki})$ inhibits all other neurons in its map $w_{ki}$ from spiking
for the remainder of the time steps of the current spiking image. Further,
these $m$ neurons can inhibit each other depending on their relative location
as we now explain. Suppose neuron $(u_{ki},v_{ki})$ of map $w_{ki}$ has the
highest potential of the $m$ neurons in (\ref{stdp1}). Then, in an
$11\times11$ area centered about $(u_{ki},v_{ki}),$ this neuron inhibits all
neurons of all the other maps in the same $11\times11$ area. Next, suppose
neuron $(u_{kj},v_{kj})$ of map $w_{kj}$ has the second highest potential of
the remaining $m-1$ neurons. If the location $(u_{kj},v_{kj})$ of this neuron
was within the $11\times11$ area centered on neuron $(u_{ki},v_{ki})$ of map
$w_{ki},$ then it is inhibited. Otherwise, this neuron at $(u_{kj},v_{kj})$
inhibits all neurons of all the other maps in a $11\times11 $ area centered on
it. This process is continued for the remaining $m-2$ neurons. In summary,
there can be no more than one neuron that spikes in the same $11\times11$ area
of all the maps.\footnote{The use of the number 11 for the $11\times11$
inhibition area of neurons was suggested by Dr. Kheradpisheh
\cite{Kheradpisheh}.} The right side of Figure \ref{l2ip2_inhbitions} shows
the final winner spike accumulation for $12$ time steps across $30$ maps 
after both lateral inhibition and STDP competition have
been imposed. It is also shows that there is at most one winner spike from all the
maps in any $11\times11$ area. For this particular input image (the number 5),
these five winner spikes are from maps 14, 16, 19, 21, and 23 at locations (19, 4),
(3,10), (17, 15), (9,12) and (3,19), respectively and will result in updates
for these 5 map kernels (weights). Lateral inhibition STDP competition resulted in an average
of only \emph{5.8 spikes per image} \ from the $30\times22\times22$ neurons in
$L2$ during training with \textsc{EMNIST} dataset. Figure \ref{l1_filter_evolution} shows how the randomly
initialized weights evolved for all 30 maps after training with 6000 images.%

%EndExpansion

\subsubsection{Spike Feature Vectors \label{spike_feature_vector}}

After (unsupervised) training of the weights (synapses) in the $L2$ layer,
these weights are fixed. Spike feature vectors are created by passing spiking
input images through layer $L2$ (Conv1) \emph{with lateral inhibition
enforced} and \emph{without STDP competition} as there is no training
involved. The spikes coming out of the $L2$ layer are then pooled in the $L3$
layer \emph{without lateral inhibition}. The pooling is done on an area of
$2\times2$ neurons in L2 with a stride of 2. Specifically, in each $2\times2$
area of L2 which contains 4 neurons, the spike of the neuron with the maximum
membrane potential $V_{L2},$ assuming it exceeds the threshold $\gamma_{L2}$, is
then the spike of the corresponding neuron of the $L3 $ (pooling) layer (i.e.,
thresholding on maxpooling). For the EMNIST dataset each input image results
in a spike tensor of shape $\tau\times30\times11\times11$. We set $\tau$ to be 
$12$ and these tensors were summed across it's first axis (i.e, along time). 
The resulting tensors in $R^{30\times11\times11}$ were flattened.\footnote{If more convolution layers
are desired, spike tensors collected in L3 layer can be used for unsupervised
training of any subsequent convolutional layers.}

Once a neuron in $L3$ spikes, it is not allowed to spike again for the 
rest of the time steps in the current image.
This results in the spike feature vectors being \emph{binary} valued (i.e.,
vectors of zeros and ones). In our experiments an average of $125$
spikes/image come out of L3 from the $30\times11\times11=3630$ neurons for the
\textsc{EMNIST} dataset. As the activations of L4 are \emph{binary} (non
differentiable), in order to do the backpropagation from layer L5 back to
layer L3 a \emph{surrogate} gradient is used (see Section \ref{surrogates} below).

\begin{figure}[!t]%
\centering
\includegraphics[
height=2.25in,
width=3.4in
]%
{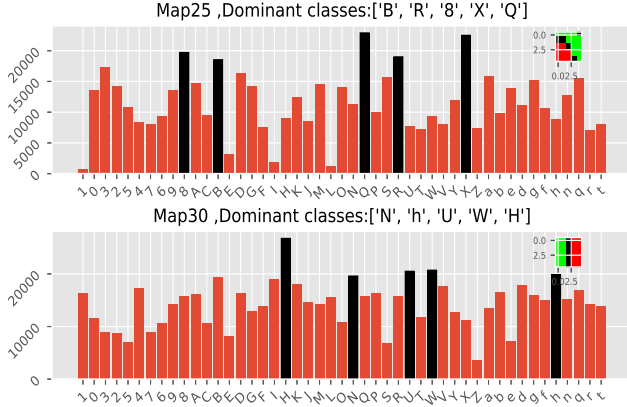}%
\caption{Spikes per map per label in L3 (Pool 1). Highlighted (in black) are the
classes that resulted in most number of spikes in a particular feature map.
Feature learned by the corresponding map is shown in the inset.}%
\label{l3_spikes_per_map_per_label_no_inh}%
\end{figure}
%EndExpansion

\subsubsection{Weight Initialization \label{weight_init}}
The weights of the $L2$ layer are initialized from the normal distribution
$\mathcal{N}(0.8,0.04)$. The weights of layers $L4$ \& $L5$ layers are
initialized from the normal distribution $\mathcal{N}(0,0.01),$ but truncated
to keep them between $\pm0.02.$ A softmax activation is used for the
classification layer $L5$ with its (net) inputs converted to integers using
the floor function. A table of values of the exponential function $e^{x}$ can
be stored in a look-up table so that the softmax activation can be calculated
using this lookup table in a hardware implementation. The activations
functions in layer $L4$ (denoted $\sigma$ in Figure \ref{network1}) are
discussed below (see Section \ref{surrogates} below).

\subsection{Spike Timing Dependent Plasticity (STDP)}

Spike timing dependent plasticity defines how a synapse (weight) between an
input (pre-synaptic) neuron and an output (post-synaptic) neuron is modulated
(updated). In its simplest form \cite{Kheradpisheh_2016}, STDP strengthens the
synapse (weight) between an input and output neuron if the input neuron aids
the output neuron in overcoming the membrane threshold (spiking); otherwise
the synapses are weakened. With $t_{out}$ and $t_{in}$ denoting the spike time
of the output (post-synaptic) and the input (pre-synaptic) neuron,
respectively, the STDP learning rule used here is given by
\begin{equation}
{\small
\begin{split}
&\text{ \ }\Delta w_{i}=\begin{cases}
-a^{-}w_{i}(1-w_{i}),\ \ \text{if}\ \ t_{out}-t_{in}<0\\
+a^{+}w_{i}(1-w_{i}),\ \ \text{if}\ \ t_{out}-t_{in}\geq0 
\end{cases}\\
&w_{i}\leftarrow w_{i}+\Delta w_{i}
\label{stdp}%
\end{split}
}
\end{equation}
Learning in spiking networks refers to the change $\Delta w_{i}$ in the
(synaptic) weight. The learning rate parameters $a^{+}$ and $a^{-}$ are
initialized with low values $(0.004,0.003)$ \cite{Kheradpisheh_2016}
\cite{Kheradpisheh_2016b} and are typically increased as the learning
progresses. In our experiments we doubled the learning rate for every $1500$ input images. As there are neither labels nor a cost function involved in the
process of STDP, it is an \emph{unsupervised} learning algorithm. That is, the
weights can be updated during the feed forward step in SNNs. In contrast, ANNs update 
their weights during error the feed back step. So, STDP does not suffer from the
update locking phenomenon \cite{frenkel2019}. Synapses in feature extraction section of the
network in Figure \ref{network1} were updated at the end of every time step.

\section{Backpropagation in the L3-L5 Layers}

Stochastic gradient descent (SGD) via backpropagation is the primary choice
for state-of-the-art classification, regression, and generative learning. A
cost function is assigned to the last layer of the network and the synapses
are updated to minimize the cost. In our network, backpropagation is used only
in the classification layers (L3-L4-L5) of the network which has a single hidden
layer L4. Let $\delta^{l},a^{l} (=\sigma(z^{l})),b^{l},W^{l},z^{l}(=w^{l}z^{l-1}+b^{l})$ denote the error vector,
the activation vector, the bias vector, the weights and the net input to the
activation function for the $l^{th}$ layer, respectively \cite{Nielsen}. $\sigma$ is the activation function. With
$C$ denoting the output cost, the backpropagation equations are
\begin{equation}
\delta^{L}=\nabla_{a}C\odot\sigma^{\prime}(z^{L}) \label{bp1a}%
\end{equation}
where $\delta^{L}$ denotes the error vector on the last layer and the error vector for the hidden layers is given by%
\begin{equation}
\delta^{l}=((W^{l+1})^{T}\delta^{l+1})\odot\sigma^{\prime}(z^{l}) \label{bp2}%
\end{equation}
Updates to biases and weights of layer $l$ are calculated with%
\begin{equation}
\frac{\partial C}{\partial b^{l}}=\delta^{l} \label{bp3}%
\end{equation}%
\begin{equation}
\frac{\partial C}{\partial W^{l}}=\delta^{l}a^{(l-1)T} \label{bp4}%
\end{equation}
$C$ denotes the cost in the final layer. We used a softmax activation with a
cross entropy cost function for the last layer so that equation (\ref{bp1a})
becomes%
\begin{equation}
\delta^{L}=-(y-a^{L}), \label{bp5}%
\end{equation}
where $a^{L}$ and $y$ are softmax activation of the output layer and the one
hot label vector, respectively.

\section{Surrogate for the Gradient\label{surrogates}}

The output activation function of $L3,L4$ layers is discontinuous and
consequently it does not have a derivative. Here we give two different
possible functions that we used to take the place of the gradient, i.e., be its surrogate \cite{courbariaux2016}.

\subsection{Surrogate Gradient 1 \label{apprx_1}}

\begin{figure}
\centering
\includegraphics[
height=2.0in,
width=3.25in
]%
{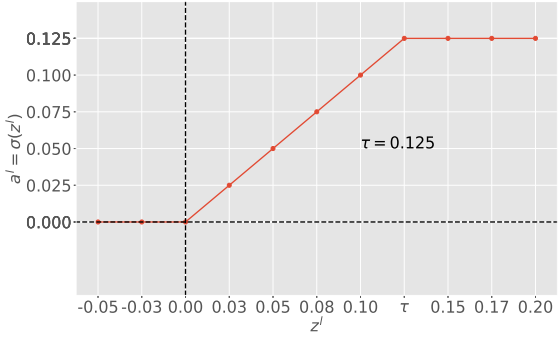}%
\caption{Activation function $a^{l}=\sigma(z^{l})$ for neurons in layer $L4.$
}%
\label{activation1}%
\end{figure}

The activation function of a neuron in layer $L4$ is defined by%
\begin{equation}
a^{l}=\sigma(z^{l})\triangleq%
\begin{cases}
0, & z<0\\
z, & 0\leq z<\tau\leq1\\
\tau, & z\geq\tau.
\end{cases}
\label{activ}%
\end{equation}
Figure \ref{activation1} is a plot of this activation function which is a ReLU
that saturates at $\tau\leq1$.

The activation activation is required to be binary so its definition is
modified to be ($\lceil\cdot\rceil$ denotes the ceiling function)%
\begin{equation}
a^{l}=\lceil\sigma(z^{l})\rceil\triangleq%
\begin{cases}
1, & z \geq 0\\
0, & z < 0%
\end{cases}
\label{app_activ}%
\end{equation}
For this activation (\ref{app_activ}) we define its surrogate gradient to be
\begin{equation}
\sigma^{\prime}(z^{l})\triangleq%
\begin{cases}
1, & 0\leq z<\tau\leq1\\
0, & \text{otherwise.}%
\end{cases}
\label{d_activ}%
\end{equation}%
which is the derivative of Equation \ref{activ}.

\begin{figure}
\centering
\includegraphics[
height=2.0in,
width=3.4in
]%
{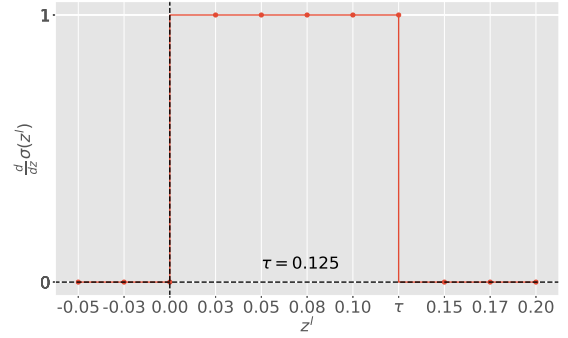}%
\caption{Surrogate gradient of activation function defined in equation
(\ref{activ}).}%
\label{d_activation1}%
\end{figure}
%EndExpansion

Simulations were performed by setting $\tau$ to $0.25,0.125,0.05$ and we found
that $0.125$ maximizes the validation accuracy. Since error backpropagation is
not feasible with equation (\ref{app_activ}), we take derivative of
$\sigma(z)$ to be equation (\ref{d_activ}). For convenience, we denote an
activation value of $1$ as spike and an activation value of $0$ as no spike.

\subsection{Surrogate Gradient 2 \label{apprx_2}}

We also considered a second activation given by%
\begin{equation}
a^{l}=\sigma(z^{l})\triangleq%
\begin{cases}
1, & z\geq0\\
0, & z<0%
\end{cases}
\label{app2_activ}%
\end{equation}
and define its surrogate gradient to be
\begin{equation}
\sigma^{\prime}(z^{l})\triangleq%
\begin{cases}
1, & z\geq0\\
0, & z<0
\end{cases}
\label{app2_d_activ}%
\end{equation}
Note that $\sigma^{\prime}(z)=\sigma(z)$ and is binary so that $a^{l}%
=\sigma^{\prime}(z^{l})$ in the hidden layer. Equation (\ref{bp2}) then
becomes
\begin{equation}
\delta^{l}=((W^{l+1})^{T}\delta^{l+1})\odot a^{l} \label{bp2a}%
\end{equation}
where $a^{l}$ determines if a neuron spikes in the $l^{th}$ layer. Hence
$a^{l}$ determines if a neuron in the $l^{th}$ layer is to receive error
information from the $l+1$ layer. Substituting Equation (\ref{bp2a}) in
Equation (\ref{bp4}) gives%
\begin{equation}
\frac{\partial C}{\partial W^{l}}=\left(  (w^{l+1})^{T}\delta^{l+1}\odot
a^{l}\right)  a^{(l-1)T} \label{bp6}%
\end{equation}
We see that a neuron in $l-1$ layer gets to update its synapse with a neuron
in $l^{th}$ layer if both neurons have spiked, i.e., for $\partial C/\partial
W_{pq}^{l}$ to be a non-zero both $a_{p}^{l}$ and $a_{q}^{l-1}$ have to be non-zero.

\section{MNIST}

\begin{figure}%
\centering
\includegraphics[
height=2.5in,
width=3.5in
]%
{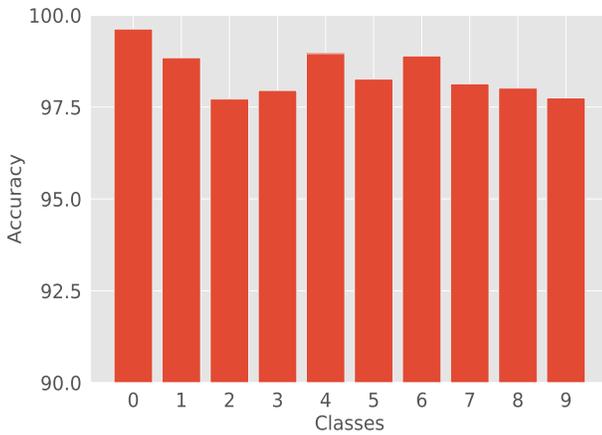}%
\caption{Classification accuracy per class of \textsc{MNIST} dataset with
surrogate gradient 1.}%
\label{mnist_acc_per_class}%
\end{figure}

Our interest here is the \textsc{EMNIST} dataset. However, as the \textsc{MNIST}
handwritten digits dataset is a popular benchmark, we briefly present our
results with it \cite{mnist}. The \textsc{MNIST} digits were passed through
the network in Figure \ref{network1} and encoded into spike vectors (described
in Section \ref{spike_feature_vector}). Note that the extracted features are
\emph{binary }valued. Table \ref{mnist_table_appx} shows that surrogate
gradient 1 yields a test accuracy $0.74\%$ higher or $74$ more correct
classifications compared to surrogate gradient 2 with $10,000$ test images. Figure
\ref{mnist_acc_per_class} shows the classification accuracy per class using
the surrogate gradient 1. 
\begin{table}
\centering%
\caption{Classification accuracies on the MNIST dataset. Dropout ($50\%$) mechanism was used
in hidden layer for regularization and number of neurons in layer L4 were set
to 900. These results were obtained by averaging over five experiments with
the classification layers of the network (in Figure \ref{network1}) trained
for $30$ epochs each time. For accuracies reported using the actual gradient a
quadratic cost function with a ReLU activation function for layers L4, L5 was used whereas for accuracies reported using the surrogate gradients a cross-entropy cost function with softmax approximation (see Section
\ref{weight_init}) for layer L5 and binary activation function for layers L3, L4 was used. Mini Batch size was set to 5. $\eta$ for the actual and surrogate gradients was set to $0.0125$ and $0.01$, respectively.}%
\begin{tabular}
[c]{|c|c|c|}\hline
Gradient Type & Mean Test Acc. & Max. Test Acc.\\\hline
Actual Gradient & $98.58\%$ & $98.66\%$ \\\hline
Surrogate Gradient 1 & $98.49\%$ & $98.54\%$ \\\hline
Surrofate Gradient 2 & $97.75\%$ & $97.77\%$ \\\hline
\end{tabular}
\label{mnist_table_appx}%
\end{table}%

\section{EMNIST}

\textsc{EMNIST} dataset has 47 classes containing handwritten upper \& lower
case letters of the English alphabet in addition to the digits. This dataset
is divided into $102,648$ training images, $10,151$ validation images, and
$18,800$ test images \cite{emnist}.

\subsection{Backpropagation with Gradient}

The features were extracted in an unsupervised fashion in layers L1, L2, and
L3 of the network (Figure \ref{network1}). As described in Section
\ref{spike_feature_vector} the neurons in L3 can spike no more than once for
an image resulting in \emph{binary} valued spike feature vectors (i.e.,
vectors of 0s and 1s). These extracted \emph{binary }valued spike feature
vectors were classified using an ANN with a ReLU activation for the hidden
layer $L4$ neurons and a softmax output activation function. The classification accuracies on EMNIST
dataset are given in Table \ref{table2}. 
\begin{table}[h]
\centering%
\caption{Classification accuracies on EMNIST dataset. Dropout of $50\%
$ was used in the hidden layer and the number of neurons in layer L4 was
1500.\textbf{\ }These results were obtained by averaging over five experiments
and trained for $25$ epochs each time. }%
\begin{tabular}
[c]{|c|c|c|c|c|}\hline
Gradient Type & Mean Test Acc. & Max Test Acc. & $\eta$ & Activation\\\hline
Actual Gradient & $85.47\%$ & $85.7\%$ & $0.05$ & ReLU\\\hline
\end{tabular}
\label{table2}%
\end{table}

\begin{figure*}[!t]%
\centering
\includegraphics[
height=1.5in,
width=7.0in
]
{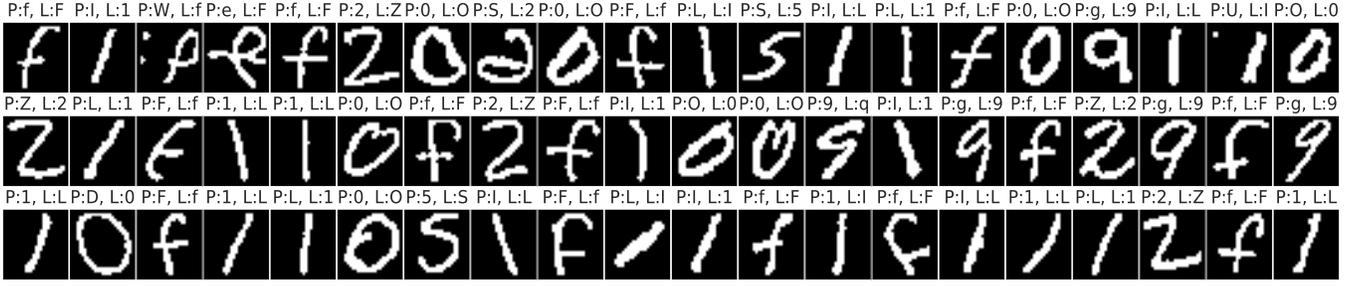}
\caption{Frequently misclassified classes in the \textsc{EMNIST} dataset.
\emph{P} and \emph{L} denote predicted class and actual label, respectively.}%
\label{misclass_regular_backprop_softmax_selected_v2_1}%
\end{figure*}

\begin{figure}
\centering
\includegraphics[
height=2.6in,
width=3.4in
]%
{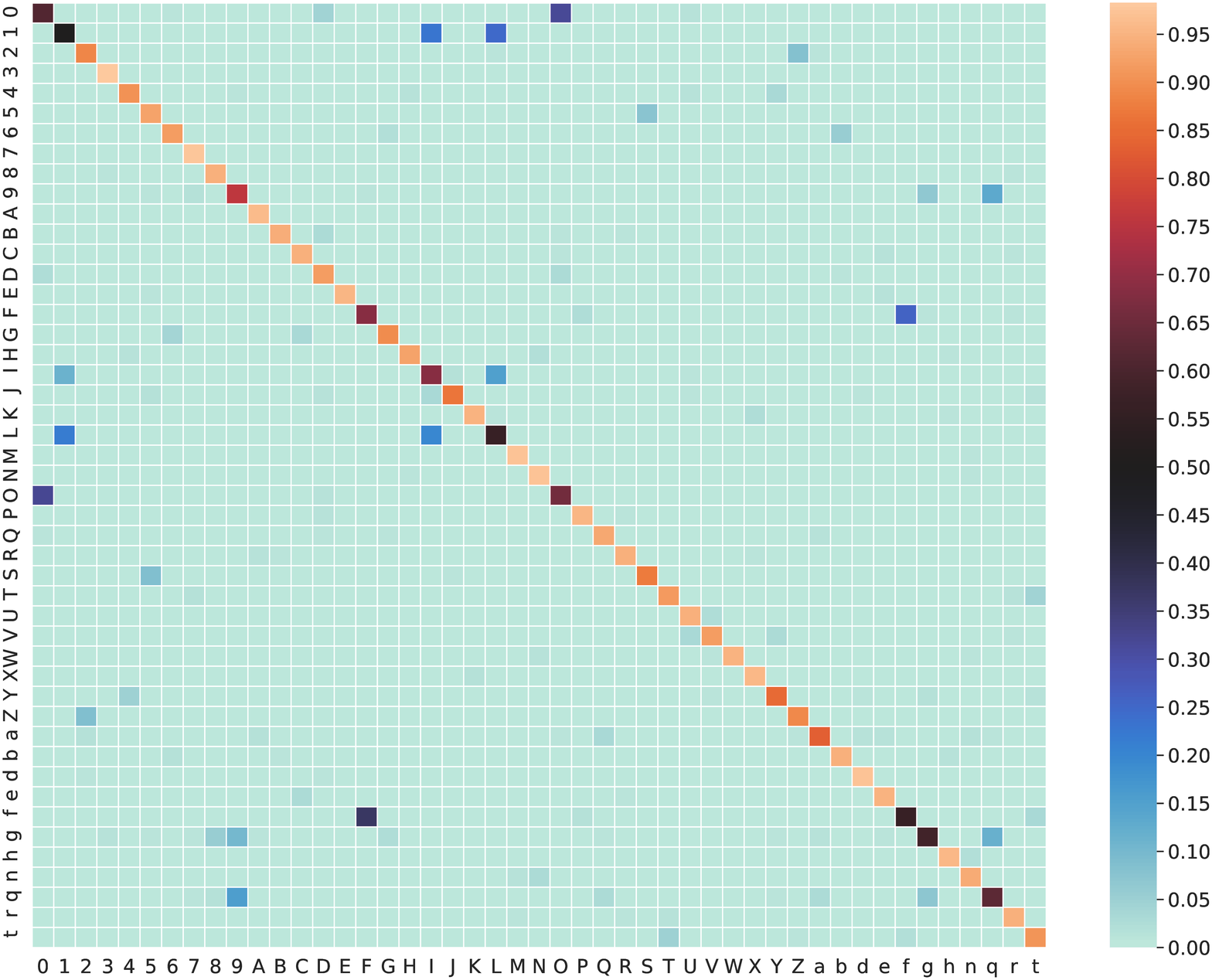}%
\caption{Confusion matrix of predictions with \textsc{EMNIST} dataset.}%
\label{heatmap_regular_backprop_final}%
\end{figure}

Figure \ref{misclass_regular_backprop_softmax_selected_v2_1} shows $60$
examples of misclassified classes. About 2688 (14.3\%) of 18800 test images were
misclassified. On further examination we found that classes \{f,F\}, \{0,O\}
the digit \textquotedblleft0\textquotedblright\ and upper case
\textquotedblleft O\textquotedblright, \{q,9\} lower case \textquotedblleft
q\textquotedblright\ and the digit \textquotedblleft9\textquotedblright, \{1,
I, L\} the digit \textquotedblleft1\textquotedblright, upper case
\textquotedblleft I\textquotedblright\ (eye) and upper case \textquotedblleft
L\textquotedblright, \{S,5\} upper case \textquotedblleft S\textquotedblright%
\ and the digit \textquotedblleft5\textquotedblright, \{2,Z\} the digit
\textquotedblleft2\textquotedblright\ and upper case \textquotedblleft
Z\textquotedblright\ were frequently misclassified. For example, in the upper
left corner of Figure \ref{misclass_regular_backprop_softmax_selected_v2_1}
the network predicted a lower case \textquotedblleft f\textquotedblright%
\ while the label was an upper case \textquotedblleft F\textquotedblright.
Figure \ref{heatmap_regular_backprop_final} shows the confusion matrix for the
classified data. For example, the digit \textquotedblleft$0$\textquotedblright%
\ was mistaken to be an upper case \textquotedblleft O\textquotedblright%
\ frequently. Similarly, upper case \textquotedblleft I\textquotedblright\ was
often mistaken to be an upper case \textquotedblleft L\textquotedblright.%

\begin{figure}
\centering
\includegraphics[
height=2.75in,
width=3.4in
]%
{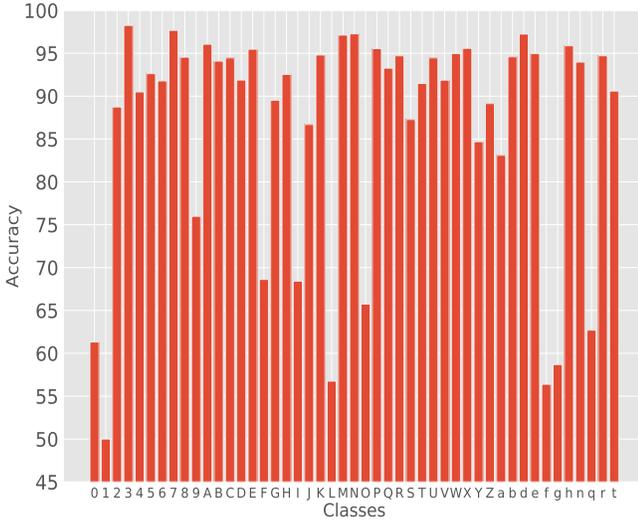}%
\caption{Accuracy per class of EMNIST dataset. For example, only about 61\% of
digits \textquotedblleft0\textquotedblright\ were classified correctly as many
of them were misclassified as the letter \textquotedblleft O".}%
\label{acc_per_class_regular_backprop_final}%
\end{figure}

%EndExpansion

\subsubsection{Conditioning on Upper Case, Lower Case and Digits}

\begin{figure}%
\centering
\includegraphics[
height=2.6in,
width=3.4in
]%
{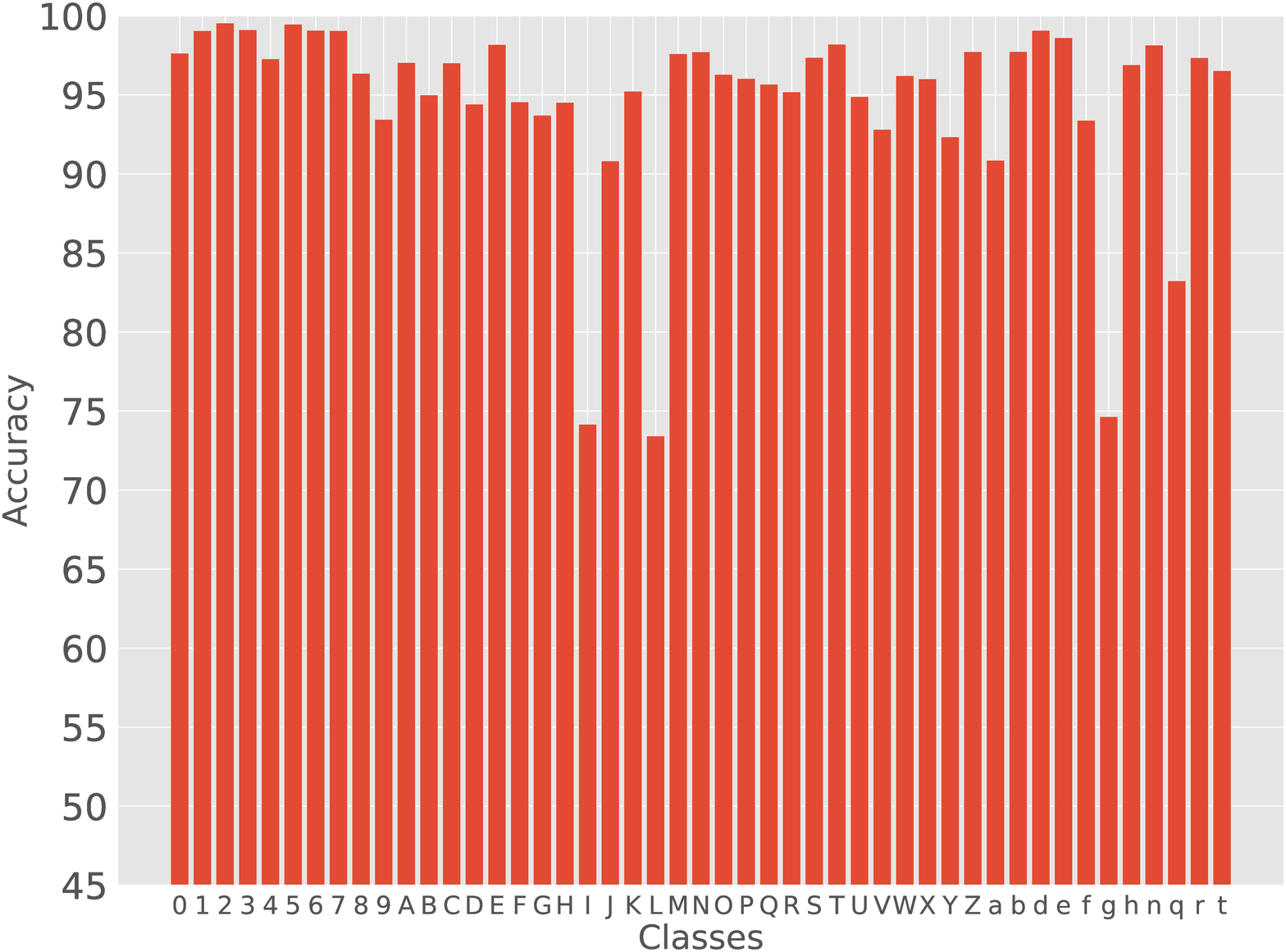}%
\caption{Accuracy per class after conditioning.}%
\label{acc_per_class_cond_regular_backprop_final}%
\end{figure}

With handwritten data, even a human classifier may not be able to tell the
difference between, for example, the upper case letter \textquotedblleft
O\textquotedblright\ and the digit \textquotedblleft0\textquotedblright. To
study this we also ran the classifier conditioned on (given that) the image
under test was an either an upper case letter, a lower case letter or a digit.
No retraining was done for this section. Table \ref{Table3} shows the dramatic
increase in accuracy from $85.6\%$ to $94.49\%$ when using this conditioning.
The accuracy per class using this conditioning is given in Figure
\ref{acc_per_class_cond_regular_backprop_final}. It is seen that the classes
I, L, g, q have the least recognition rate, but still well above their
accuracies given previously in Figure
\ref{acc_per_class_regular_backprop_final} where conditioning was not used. In
more detail we found that about $11\%$ of the letters \textquotedblleft
q\textquotedblright\ were misclassified as the letter "g", about $4\%$ of
letters \textquotedblleft q\textquotedblright\ were misclassified as the
letter \textquotedblleft a\textquotedblright, while about $84\%$ of letters
"q" were correctly classified. About $26\%$ of letters \textquotedblleft
g\textquotedblright\ were misclassified as the letter \textquotedblleft
q\textquotedblright\ while about $67\%$ of letters \textquotedblleft
g\textquotedblright\ were correctly classified. Similarly, we found that about
$22\%$ of letters of upper case \textquotedblleft I\textquotedblright\ (eye)
were misclassified as the upper case letter \textquotedblleft L" while $73\%$
of upper case \textquotedblleft I\textquotedblright\ were correctly
classified. As a final observation about $22\%$ of upper case letters
\textquotedblleft L\textquotedblright\ were misclassified as an upper case
\textquotedblleft I\textquotedblright\ (eye) while about $76\%$ of upper case
letters \textquotedblleft L\textquotedblright\ were correctly classified.
\begin{table}[h]
\centering%
\caption{Classification accuracies on EMNIST dataset conditioned on input
being a digit, upper case or a lower case letter. }%
\begin{tabular}
[c]{|c|}\hline
Conditioned Maximum Test Accuracy\\\hline
94.49 \%\\\hline
\end{tabular}
\label{Table3}%
\end{table}%

\begin{figure}%
\centering
\includegraphics[
height=2.6in,
width=3.4in
]%
{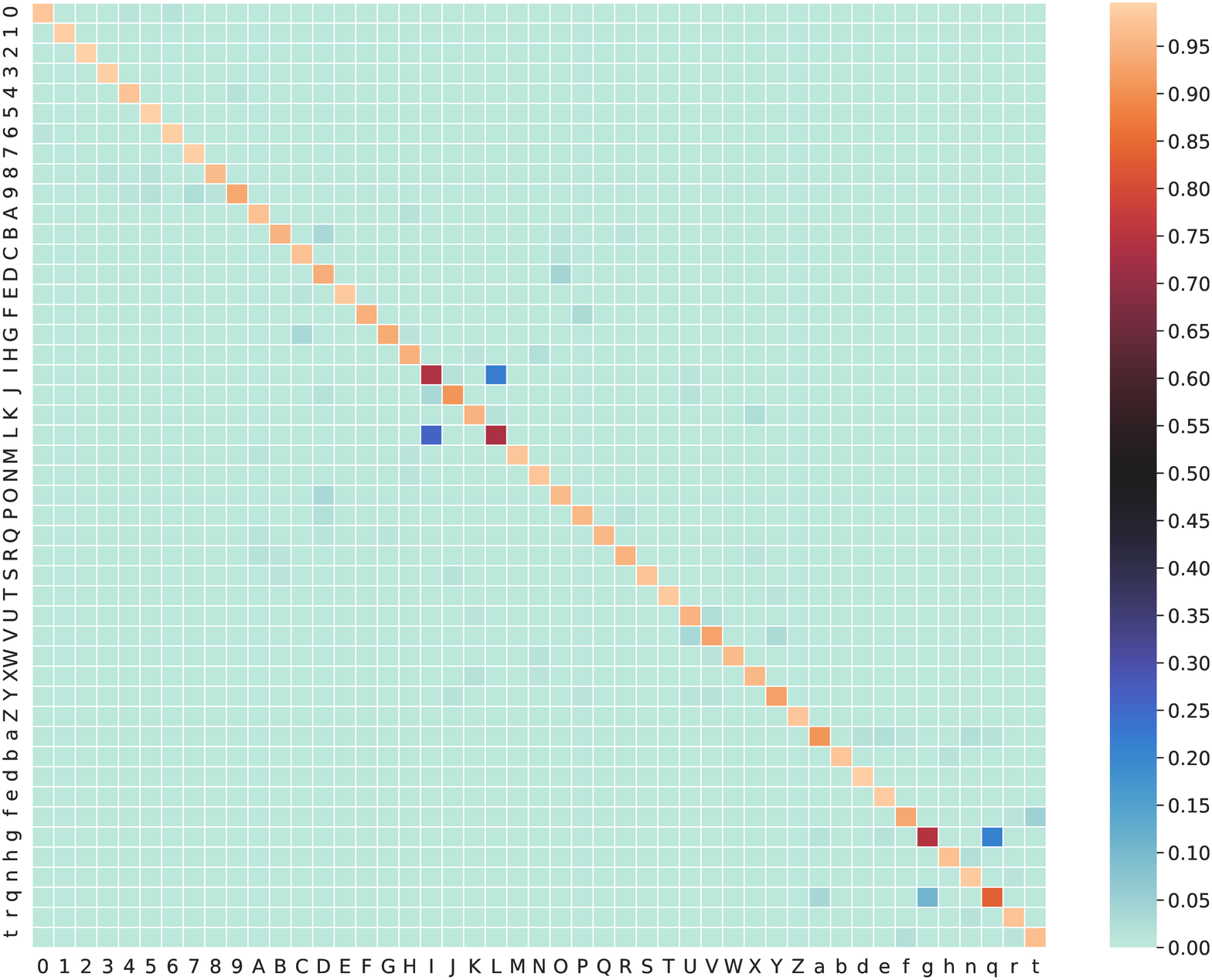}%
\caption{Confusion matrix of classification with \textsc{EMNIST} dataset after
the input was conditioned on being a digit, upper case or a lower case letter.}%
\label{heatmap_cond_regular_backprop_final}%
\end{figure}

\subsection{Backpropagation with Gradient Surrogates}

In this section \emph{binary} valued features vectors (i.e., vector with 0s
and 1s) were collected in layer $L3$ as described in Section
\ref{spike_feature_vector}. Classification was performed using an ANN with
binary activation for the hidden layer $L4$ neurons and an approximated
softmax output explained in Section \ref{weight_init}.

\begin{table*}[t]
\centering%
\caption{Classification accuracies on EMNIST dataset. Dropout ($50\%$) mechanism was
used in hidden layer for regularization and number of neurons in L4 was set to
1500. A cross entropy cost function with softmax approximation (see Section
\ref{weight_init}) was used. These results were obtained by averaging over five experiments with the
classification layers of the network (in Figure \ref{network1}) trained for
$25$ epochs each time. Batch size was set to 5.}%
\begin{tabular}
[c]{|c|c|c|c|c|c|}\hline
Gradient Type & Mean Test Acc. & Max. Test Acc. & Conditioned Max. Test Acc. &
$\eta$ & Activation\\\hline
Surrogate Gradient 1 & 85.35 \% & 85.49 \% & 94.1 \% & 0.02 & Binary\\\hline
Surrogate Gradient 2 & 84.24 \% & 84.47 \% & 93.72 \% & 0.02 & Binary\\\hline
\end{tabular}
\label{emnist_table_apprx}%
\end{table*}

\noindent Table \ref{emnist_table_apprx} shows that the gradient surrogate 1
outperforms gradient surrogate 2 by $1.0\%$ ($188$ more correct
classifications with 18800 test images).

\paragraph{Computational Advantage of Binary Activations}

In the feedforward paths L1 through L4 the matrix-vector multiplication
operations can all be avoided in a hardware implementation as these layers all
have binary activations. For example, executing the multiplication of a set of
(floating point) weights times a set of spikes (binary activations) is simply
\begin{equation}
{\small
{\displaystyle\left(  {%
\begin{array}
[c]{ccc}%
\text{\textrm{w}}_{11} & \mathrm{w}_{12} & \mathrm{w}_{13}\\
\mathrm{w}_{21} & \mathrm{w}_{22} & \mathrm{w}_{23}\\
\mathrm{w}_{31} & \mathrm{w}_{32} & \mathrm{w}_{33}\\
& \vdots & \\
\mathrm{w}_{n1} & \mathrm{w}_{n2} & \mathrm{w}_{n3}%
\end{array}
}\right)  \times\left(  {%
\begin{array}
[c]{c}%
\mathrm{0}\\
\mathrm{1}\\
\mathrm{1}%
\end{array}
}\right)  =\left(  {%
\begin{array}
[c]{c}%
\mathrm{w}_{12}\\
\mathrm{w}_{22}\\
\mathrm{w}_{32}\\
\vdots\\
\mathrm{w}_{n2}%
\end{array}
}\right)  +\left(  {%
\begin{array}
[c]{c}%
\mathrm{w}_{13}\\
\mathrm{w}_{23}\\
\mathrm{w}_{33}\\
\vdots\\
\mathrm{w}_{n3}%
\end{array}
}\right)  .} \label{feed_fwd}%
}
\end{equation}
That is, multiplication is replaced by addition. This technique avoids the
need for dedicated multiplier hardware and allows the feasibility of in memory
computing \cite{wu}\cite{wu2015}

Another advantage is found in backpropagation computations. Specifically, as
the surrogate gradient $\sigma^{\prime}(z^{l})$ is binary, the error vector
$\delta^{l}$ for the hidden layer can be obtained without having to do some of
the row-column multiplications in
\[
((W^{l+1})^{T}\delta^{l+1})\odot\sigma^{\prime}(z^{l}).
\]
For example%
\begin{equation}
{\small
{\displaystyle}\underset{(w^{l+1})^{T}}{\underbrace{{\left(  {%
\begin{array}
[c]{cc}
\mathrm{w}_{11} & \mathrm{w}_{12} \\
\mathrm{w}_{21} & \mathrm{w}_{22}  
\end{array}
}\right)  }}}{\times}\underset{\delta^{l+1}}{\underbrace{{\left(  {%
\begin{array}
[c]{c}
\mathrm{1}\\
\text{\textrm{2.5}}\\
\end{array}
}\right)  }}}{\odot}\underset{\sigma^{\prime}(z^{l})}{\underbrace{{\left(
\begin{array}
[c]{c}%
0\\
1
\end{array}
\right)  }}}{\\
=\left(
\begin{array}
[c]{c}%
0\\
\mathrm{w}_{21}+2.5\mathrm{w}_{22}
\end{array}
\right)  } \label{del_l}%
}
\end{equation}

That is, in equation (\ref{del_l}) the row-column multiplications of the first
row are avoided as the result will zero due to the element-wise (Hadamard
product) vector multiplication. All the weight updates, $\partial C/\partial
W^{l}$ can be obtained without explicitly calculating vector outer product
$\delta^{l}a^{(l-1)T}$ as the activations of $L3$ and $L4$ layers are
binarized. For example%
\begin{equation}
{\displaystyle}\underset{\delta^{l}}{\underbrace{{\left(  {%
\begin{array}
[c]{c}%
\mathrm{a}\\
\mathrm{b}\\
\mathrm{c}%
\end{array}
}\right)  }}}{\times}\underset{a^{(l-1)T}}{\underbrace{{\left(  {%
\begin{array}
[c]{ccc}%
0 & 1 & 0
\end{array}
}\right)  }}}{=\left(  {%
\begin{array}
[c]{ccc}%
0 & a & 0\\
0 & b & 0\\
0 & c & 0
\end{array}
}\right)  .} \label{dc_dw}%
\end{equation}
That is, the matrix on the right side of Equation (\ref{dc_dw}) is found by
simply transcribing $\delta^{l}$ into its columns as specified by $a^{(l-1)T}$.

\section{Software Tool}
\begin{figure}%
\centering
\includegraphics[
height=1.5in,
width=3.3in
]%
{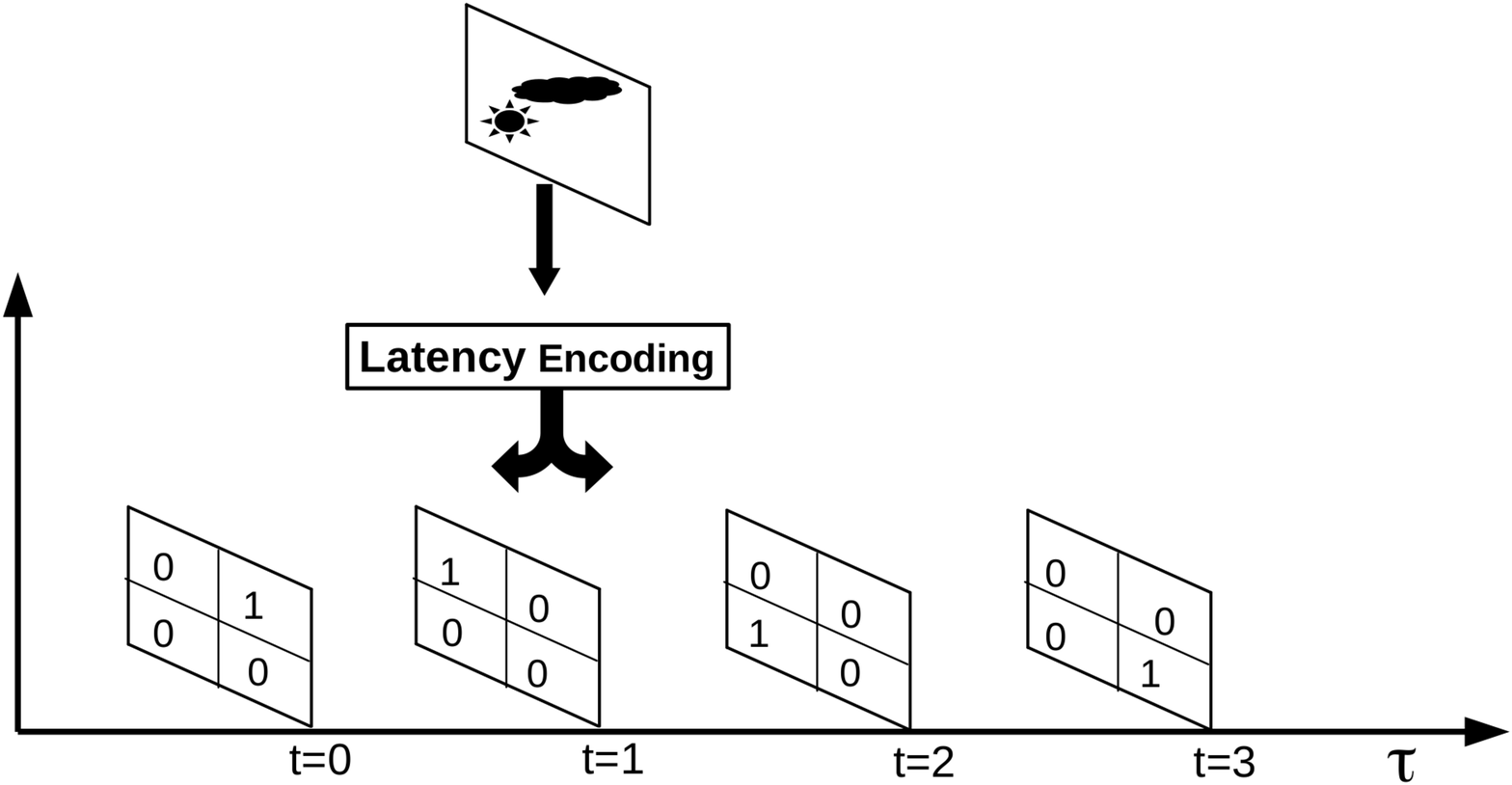}%
\caption{Spikes are represented with a value 1 and no spike is represented
with a value 0. In this figure $\tau$ was set to 4.}%
\label{roc_encoding}%
\end{figure}

Previously the authors have used the \textsc{PyNN} simulator with
\textsc{Neuron} \cite{vaila2018} \cite{neuron}. However these are tools for
neuroscientists with neuron models much more complex than needed in our case.
Even tools like \textsc{Nengo} \cite{nengo} (developed for bio-inspired
machine learning) use more complex neuronal models than necessary here.
Motivated by the simple spiking models in Kheradpisheh et al.'s work in
\cite{Kheradpisheh_2016}, we developed our software tools. Following
\cite{Kheradpisheh_2016} our package supports instantaneous (non leaky
integrate and fire) neurons, latency encoding, and inhibition mechanisms to be
able to simply extract meaningful features from the input images. Feature
extraction in SNNs is unsupervised in contrast to ANNs. To monitor the weight
updates (synapse changes) in the spiking network, the software provides the
capability to monitor spike activity, weight evolution (updates), feature
extraction (spikes per map per label), and synapse convergence, etc. This
software tool was used here and in \cite{vaila2019} \cite{vaila2019a}. Similar
to our work, Mozafari et al released the software tool \textsc{SpykeTorch} in
\cite{spyketorch} which is based on the \textsc{PyTorch} \cite{pytorch} deep
learning tool. Our software is named \textsc{SpykeFlow}\footnote{\url{https://github.com/ruthvik92/SpykeFlow}}
 and primarily uses \textsc{NumPy} \cite{numpy} to do the calculations of
lateral inhibition, STDP updates, neuron spike accumulation, etc. However, we
also use \textsc{tensorflow} \cite{tensorflow2015} for computationally
intensive calculations such as convolution and pooling. Therefore, the users
will have the ability to use a GPU, if one is available. Visualizations are
performed using \textsc{Matplotlib} \cite{Matplotlib} and we also provide some
miscellaneous \textsc{Jupyter} notebooks. \textsc{SpykeFlow} is divided into
two main classes: feature extraction and feature classification. Firstly,
input images (2D) are converted to tensors of three dimensions with time $t$
as the extra dimension. For example, Figure \ref{roc_encoding} shows that an
input (2D) image converted to a spiking image represented by a rank 3 tensor
of binary values. Spike are arranged into $\tau$ slices based on latency
encoding described in Section \ref{network_desc}. Figure \ref{roc_encoding}
also indicates that if an input neuron spikes (binary 1) it is not allowed to
spike for the rest of the image. In all of our experiments we set $\tau=12$ so
$t=0,1,...11$. Batch size in our software is 1 time step which means the
synapses are updated after every 1 time step. The file structure of  
\textsc{SpykeFlow} is shown below

\textbf{AllDataSets/}

\textbf{spykeflow/}

\texttt{\hspace{0.2in}network.py}

\texttt{\hspace{0.2in}inputlayerclass.py}

\texttt{\hspace{0.2in}classifierclass.py}

\texttt{\hspace{0.2in}.}

\texttt{\hspace{0.2in}.}

\textbf{main/}

\texttt{\hspace{0.2in}main.py}

\textbf{notebooks/}

\textbf{outputs/}

\noindent The \texttt{AllDataSets} folder contains all the datasets that we
intend to work with and the folder \texttt{spykeflow} contains core classes of
the tool. The \texttt{main} folder contains the \texttt{main.py} file which
contains the code to perform feature extraction, visualization, and
classification. The \texttt{notebooks} folder contains the \ \textsc{Jupyter}
notebooks for classification part of the network (in Figure \ref{network1})
and show the code for various backpropagation approximations that were used
in this work. The \texttt{outputs} folder contains plots generated by
\textsc{SpykeFlow} for various datasets. Contents of \texttt{main.py} file 
are discussed below

\subsection{Conv1 and Pool 1 layers}
{\small
\begin{python}
import numpy as np
import os, sys, random  
sys.path.insert(0, "path_to_SpykeFlow")
import spykeflow as sf
os.environ["CUDA_VISIBLE_DEVICES"]="-1"
from spykeflow import network as network
from spykeflow import classifierclass as cls
from spykeflow import inputlayerclass as inputlayer
# Input layer
firstLayer = inputlayer.InputLayer(debug=False,
size=27, dataset=data_set, off_threshold=50, 
on_threshold=50, border_size=2, data='test',
val_frac=0.14,test_frac=0.14)
test_input_data=firstLayer.EncodedData()   
#returned data is in [(data_tensor1, data_label1),\
#(data_tensor2, data_label2), .......]
random.shuffle(train_input_data)
class_labels_train = map(lambda x: np.where(\
x[1]==1)[1][0], train_input_data)
labels_dict = {0:'0', 1:'1', ..,45:'r',46't'}
nofImages = 6000
train_input_images = [items[0] for items \
in train_input_data][0:nofImages]
train_input_images = np.concatenate(\
train_input_data, axis=3)
size = train_input_images.shape[0]
T = train_input_images.shape[-1] #Total time steps
#First Conv layer, pool is disabled as train=True
net1 = network.Network(output_channels=30, inputs=\
train_input_images, A_plus=0.002, debug=False,\
sample_interval=200, train=True,\
save_pool_spike_tensor=False, threshold=15.0,\
size=size, inh_reg=11)
#Start training
net1.feedforward()
\end{python}
\label{code1}
}
\noindent Running \texttt{net1.feedforward()} executes the following algorithm.

\noindent\rule{3.4in}{1pt}

\texttt{Method net1.feedforward()}

\noindent\rule{3.4in}{1pt}

$\mathtt{\hspace{0.2in}}$\texttt{for t in range(0,T)}

$\mathtt{\hspace{0.2in}\hspace{0.2in}}$\texttt{if(t\%}$\tau$\texttt{==0)\#end
of current image}

$\mathtt{\hspace{0.2in}\hspace{0.2in}\hspace{0.2in}}$\texttt{Reset neurons}

$\mathtt{\hspace{0.2in}\hspace{0.2in}}$\texttt{Feed forward for 1 timestep}

$\mathtt{\hspace{0.2in}\hspace{0.2in}}$\texttt{Lateral inhibition}

$\mathtt{\hspace{0.2in}\hspace{0.2in}}$\texttt{STDP competition}

$\mathtt{\hspace{0.2in}\hspace{0.2in}}$\texttt{Determine final spikes}

$\mathtt{\hspace{0.2in}\hspace{0.2in}}$\texttt{STDP weight updates}

$\mathtt{\hspace{0.2in}\hspace{0.2in}}$\texttt{Record weights,spikes}

\noindent\rule{3.4in}{1pt}

\begin{figure}[H]%
\centering
\includegraphics[
height=1.4in,
width=3.4in
]%
{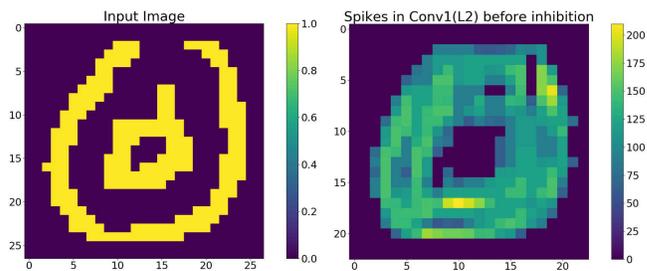}%
\caption{Left: Input spikes. Right: Neurons in in Conv1 (L2) that crossed the
threshold. Number of times a neuron in an X-Y location across all the maps
(features) that crossed the threshold is indicated by the color code. Spikes
are summed across time and number of maps for visualization.}%
\label{input_b4_inh}%
\end{figure}

\begin{figure}[H]%
\centering
\includegraphics[
height=1.4in,
width=3.4in
]%
{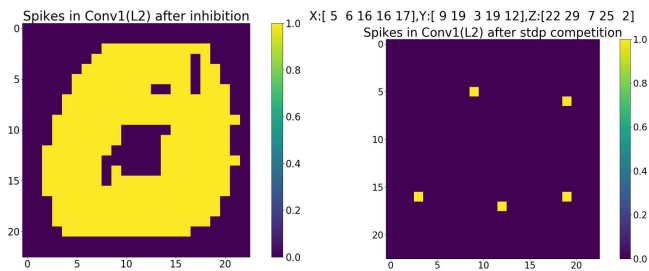}%
\caption{Left: Neurons that crossed threshold in L2 after lateral inhibition.
Right: Final spikes in L2 after STDP competition. Number of neurons that
crossed the threshold in an X-Y location is indicated by the color code. Z
indicates the map number. Spikes are summed across time and number of maps for
visualization.}%
\label{lateral_stdp}%
\end{figure}

\noindent If \texttt{debug} is set to \texttt{True} in the object
\texttt{net1} then a series of images showing the internal activity of the
network are shown. Figure \ref{input_b4_inh} and Figure \ref{lateral_stdp}
show the steps carried out in L2 (Conv1) layer to determine the spikes that
result in a weight update.%

\subsubsection{Generating plots}
Once the training of the first convolutional layer is finished, plots can be 
generated with the following code snippet for further analysis
{\small
\begin{python}
net1.feature_visualization([net1.evol_weights], 
sample_interval, intervals, plotx=5, ploty=6)
net1.animation([net1.evol_weights], plotx=5, 
ploty=6, sample_interval=sample_interval, 
intervals=intervals)
net1.spike_statistics()
\end{python}
}
\begin{figure}
\centering
\includegraphics[
height=2.0in,
width=3.4in
]%
{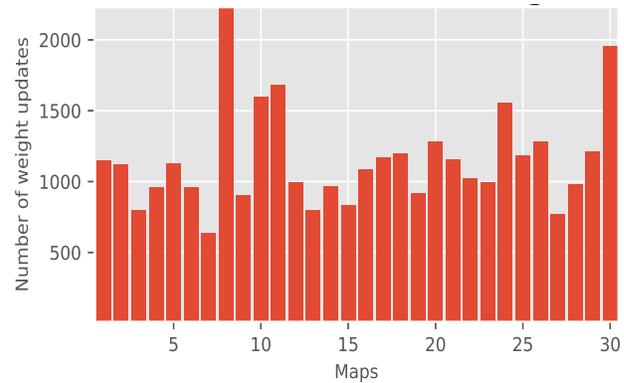}%
\caption{Number of spikes that resulted in STDP weight updates per map.}%
\label{l2_barchart_ieee}%
\end{figure}

\begin{figure}
\centering
\includegraphics[
height=2.0in,
width=3.4in
]%
{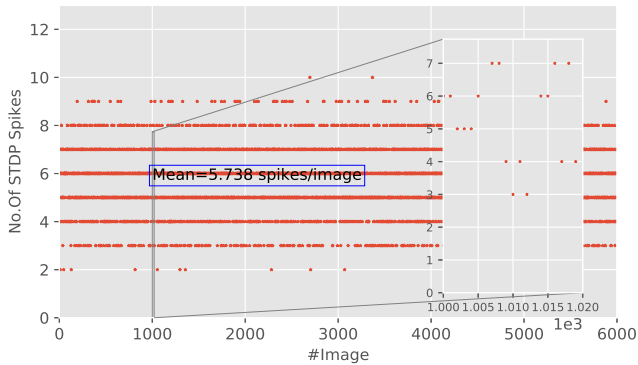}%
\caption{Spikes per image in layer L2. Shown in the inset zoomed to show images
1000 to 1020.}%
\label{l2_spikes_per_image}%
\end{figure}

The \texttt{net1.feature\_visualization} method generates a plot
given in Figure \ref{l1_filter_evolution} and the method \texttt{net1.animation}
generates an animation of evolved features in Conv1\footnote{\url{https://www.youtube.com/watch?v=KA4IJe2AtvE}}.
The \texttt{net1.spike\_statistics} method produces the
plots shown in Figures \ref{l2_barchart_ieee} and \ref{l2_spikes_per_image}. 
The first convolution layer (Conv1) was trained by passing 6000 \textsc{EMNIST} images
through the network. Figure
\ref{l2_barchart_ieee} shows the number of times each of the 30 maps had their
weights updated as the 6000 training images passed through the network. The number of STDP spikes (weight updates) for each of the 6000 training
images are shown in Figure \ref{l2_spikes_per_image}. For example, images 1000
and 1001 (see inset) each had 6 STDP\ spikes meaning that 6 of the 30 maps of
L2 had their weights updated as these images went through the network.

\subsubsection{Stopping criteria}

{\small
\begin{equation}
C_{l} \triangleq \frac{\sum\nolimits_{w,k,i,j}\left(  W_{l}(w,k,i,j)\ast(1-W_{l}%
(w,k,i,j))%
\genfrac{}{}{0pt}{}{{}}{{}}%
\right)}{n_{l}} \label{converg}%
\end{equation}
}
$W_{l}$ are the weights (synapses) of $l^{th}$ layer,
specifically, for the first convolution layer with $30$ maps, $W_{l=2}(w,k,i,j)\in%
\mathbb{R}
^{30\times2\times5\times5}$ with $n_{l=2}=30\times2\times5\times5=1500, $ 
and for the second convolution layer with $200$ maps, $W_{l=4}(w,k,i,j)\in \mathbb{R}
^{200\times30\times5\times5}$ with $n_{l=4}=200\times30\times5\times5=150000.$
Kheradpisheh et al \cite{Kheradpisheh_2016b} defined a convergence factor
given in Equation (\ref{converg}) to stop the training if $0.01<C_{l}<0.02$.
If the elements in $W_{l}$ all approach either 0 or 1 then $C_{l}$
$\rightarrow0$. This stopping criteria indicates the weights are near
saturation and no additional weight updates will then matter. In this work, 
during training, the weights (synapses) are sampled
after every 200 images (see Figure \ref{l2l4_weight_convergence_no_inh}) have passed through
the network. In the initial phases of the training, the weights (synapses)
approach 0.5 because, as Figure \ref{l2l4_weight_convergence_no_inh} indicates, the value of
$C_{l}$ approaches 0.2 (See \footnote{The weights in feature extraction layers are
bounded between 0 and 1.The maximum of $w(1-w)$ is at $w=0.25.$ $C_{l}$ is the
average of these values for all the weights and $C_{l}=0.25$ if and only if
all the weights equal 0.5.}). Another way to determine if the network is finished learning is to look at the
temporal difference of the weights (synapses) as defined by Equation
(\ref{td}). The corresponding plot for our experiment is shown in Figure
\ref{l2l4_weight_difference_no_inh}.%
\begin{equation}
{\small
\begin{split}
&\text{Temporal Difference} \triangleq \\
&\frac{\sum\nolimits_{w,k,i,j}\left(  W_{l}%
^{[t-1]}(w,k,i,j)-W_{l}^{[t]}(w,k,i,j)\right)  ^{2}.}{n_{l}} \label{td}%
\end{split}
}
\end{equation}%
\subsubsection{Collecting spikes in Pool 1 (with and without lateral inhibition)}
Once the first convolutional layer (L2) is trained, the weights of
L2 (Conv1) are fixed and input spikes from L1 are simply passed through L2 and
pooled in L3 (Pool 1). Lateral inhibition is still applied as the
spikes pass through L2 so that only the dominant feature map will be allowed
to produce spike. However, STDP competition is \emph{not} applied because there is no
training involved as the L2 weights are now fixed. There is also an option to
enforce or not to enforce lateral inhibition in L3 (Pool 1) layer. In the
following code snippet, setting \texttt{pool\_lateral\_inh=False} results in
lateral inhibition in L3 (Pool 1) being turned off. Setting
\texttt{pool\_spike\_accum=False} (as done in this work) restricts the number
of spikes per neuron in L3 to at most one. Below code snippet collects spikes
in Pool 1  \emph{without} lateral inhibition.

{\small
\begin{python}
evolved_weights = net1.evol_weights[-1]
#First Pool and Conv layers, note that train=False
net2 = network.Network(output_channels=30,\
pool_lateral_inh=False, inputs=train_input_images,\
train=False, set_weights=evolved_weights,\
debug=False, save_pool_spike_tensor=True,\
threshold=15.0, save_pool_features=True,\
pool_spike_accum=False)
net2.rewire_weights() # fixing the weights
net2.feedforward()
fig = net2.spikes_per_map_per_class(plot_x=1,
plot_y=2, class_labels=class_labels_train,\
pool_output_data=net2.pool_spike_tensor,\
labels_map=labels_map, view_maps=[25, 30],\
final_weights=evolved_weights,\
labels_map=labels_dict)
\end{python}
}

\begin{figure}[!t]%
\centering
\includegraphics[
height=2.25in,
width=3.4in
]%
{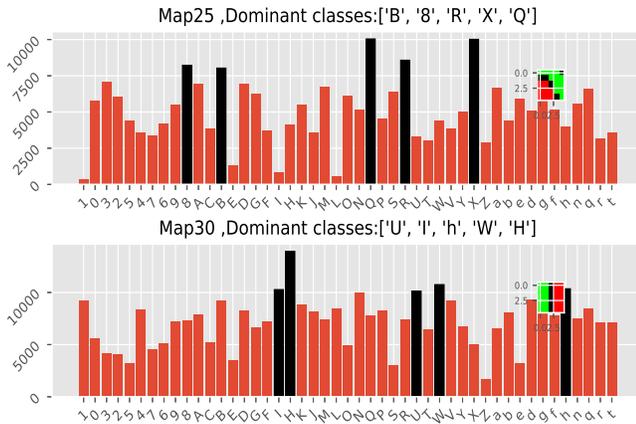}%
\caption{Spikes per map per label. Highlighted (in black) are the classes that
resulted in most number of spikes in a particular feature map. Features
learned by a map are shown in the inset. Notice that the number of spikes in
general is less compared to Figure \ref{l3_spikes_per_map_per_label_no_inh}.}%
\label{l3_spikes_per_map_per_label_inh}%
\end{figure}

If \texttt{pool\_lateral\_inh} is set to \texttt{True} then lateral
inhibition is turned on in the pooling layer. (As shown in Figure
\ref{l3_spikes_per_map_per_label_inh}, this case results in lesser spikes per
map per label compared to that of Figure
\ref{l3_spikes_per_map_per_label_no_inh}.)
Below code snippet collects spikes
in Pool 1 \emph{with} lateral inhibition.

{\small
\begin{python}
evolved_weights = net1.evol_weights[-1]
#First Pool and Conv layers, note that train=False
net3 = network.Network(output_channels=30,\
pool_lateral_inh=True, inputs=train_input_images,\
train=False, set_weights=evolved_weights,\
debug=False, save_pool_spike_tensor=True,\
threshold=15.0, save_pool_features=True,\
pool_spike_accum=False)
net3.rewire_weights() # fixing the weights
net3.feedforward()
fig = net2.spikes_per_map_per_class(plot_x=1,
plot_y=2, class_labels=class_labels_train,\
pool_output_data=net2.pool_spike_tensor,\
labels_map=labels_map, view_maps=[25, 30],\
final_weights=evolved_weights,\
labels_map=labels_dict)
\end{python}
}

\subsubsection{Feature classification in Pool 1}
The \texttt{net2.spikes\_per\_map\_per\_class} method was used to
produce Figure \ref{l3_spikes_per_map_per_label_no_inh}. This figure shows
only two of 30 maps in layers L2 and L3 as \texttt{view\_maps=[25,30]}.
The spikes for each image are collected in layer L3 as a tensor
giving $S_{L3}(t)\in%
\mathbb{R}
^{12\times30\times11\times11}.$ 12 is the number of time steps per image, 30
is the number of maps in L2/L3, and $11\times11$ is the shape of each map in
L3 (pooling layer). This spike tensor is summed over the 12 time steps giving
$S_{L3}^{total}\triangleq\sum_{t=0}^{11}S_{L3}(t)\in%
\mathbb{R}
^{30\times11\times11}.$ $S_{L3}^{total}$ is then flattened to obtain a binary
spike feature vector in $%
\mathbb{R}
^{3630}$ ($30\times11\times11=3630$). Note that the keyword
\texttt{save\_pool\_features }in the object \texttt{net2} must be set to
\texttt{True} in order to create spike feature vectors. Generated spike 
features are classified using an ANN provided in \texttt{classifierclass.py} as shown in the
below code snippet. 
However, in this work we used ANNs with binary activations and surrogate 
gradients to classify the spike feature vectors.

\begin{python}
train_pool_spike_features = net2.make_feature_vecs\
(net2.pool_spike_features)
n_classes = 47
n_hidden = 1
net_struct = \
[train_pool1_spike_features.shape[1],1500,n_classes]
log_path = 'path' 
#path to log data to be visualized in tensorboard
neural_net = cls.Classifier(train_data=\
(train_pool1_spike_features, class_labels_train),\
test_data=(test_pool1_spike_features,\
class_labels_test), network_structure=net_struct,\
activation_fns=activation_fns, epochs=10,eta=0.001,\
lmbda=0.0001, verbose=1, plots=True,\
optimizer='adam', eta_decay_factor=1.007,\
patience=8, eta_drop_type='plateau', epochs_drop=1,\
val_frac=0.091, drop_out=0.0, ip_lyr_drop_out=0.0,\
leaky_alpha=0.1, leaky_relu=False,\
weight_init='he_uniform', bias_init=0.1,
batch_size=5, log_path=log_path)
neural_net.keras_fcn_classifier()
\end{python}
\subsection{Conv2 and Pool 2 without lateral inhibition in Pool 1}
If a second convolution layer is added to the network, the
accumulated spikes in L3 (Pool 1 with or without lateral inhibitions) can be used
as input to train this second convolution layer (Conv2) as shown below. Using the spikes
collected from L3 (Pool 1) without lateral inhibition the code for this 2nd
convolution layer is as follows:

\begin{python}
nTrain_images = 35000, l4_maps = 200 
#35k are enough if lateral inh in pool1 is False
size = net2.pool_spike_tensor.shape[0]
input_channels = net2.pool_spike_tensor.shape[2]
inputs = \
net2.pool_spike_tensor[:,:,:,0:nTrain_images*tsteps]
net4 = network.Network(pool_lateral_inh=False,\
inputs=inputs, A_plus=0.0002, debug=False,\
output_channels=l4_maps, size=size,\
input_channels=input_channels, lr_inc_rate=1500,\
sample_interval=200, train=True,\
threshold=15.0, inh_reg=3, epochs=1)
net4.feedforward()
\end{python}

\subsubsection{Generating plots}
Various plots for Conv2 (Pool 2) can be generated with the following code for further analysis
\begin{python}
layer_num = [2, 3, 4] #conv1, pool1, pool2
filter_sizes = [net2.conv_kernel_size,\
net3.pool_kernel_size, net4.conv_kernel_size] 
filter_strides = [1, 2, 1]
nof_filters = [net2.output_channels,\
net3.output_channels,net4.output_channels] 
#[list of #filters from first conv to last layers]
types = ['conv', 'pool', 'conv']
layer_weights=[[net2.evol_weights],\
[net4.evol_weights]], currLayer= 4
#[list of synapses from  first conv to last conv]
fig = net4.feature_visualization(layer_weights,\
sample_interval, intervals, plotx=5, ploty=5,\
layer_num=layer_num, filter_sizes=filter_sizes,\
nof_filters=nof_filters, types=types,\
currLayer=currLayer, show=True)
net4.feature_convergence([net1.evol_weights, \
net4.evol_weights], sample_interval)
animation, fig = net4.animation(layer_weights,\
sample_interval,intervals,plotx=10,ploty=10,\
layer_num=layer_num, filter_sizes=filter_sizes,\
filter_strides=filter_strides, currLayer=currLayer,\
types=types, nof_filters=nof_filters)
\end{python}

\begin{figure}
\centering
\includegraphics[
height=1.8in,
width=3.4in
]%
{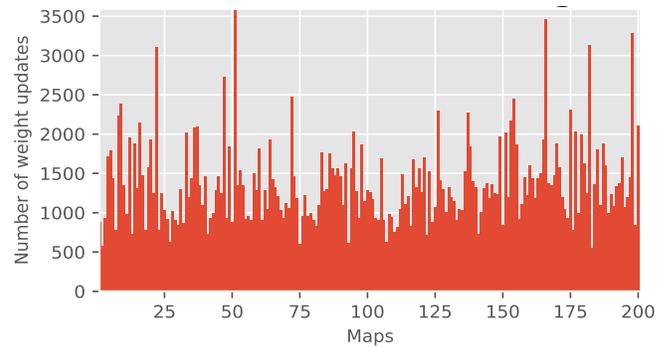}%
\caption{Number of spikes that resulted in STDP weight updates per map in Conv2.}%
\label{l4_bar_chart_weight_updates_no_inh}%
\end{figure}

\begin{figure}
\centering
\includegraphics[
height=1.8in,
width=3.4in
]%
{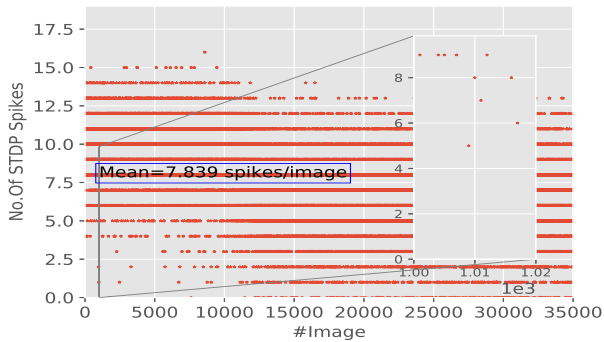}%
\caption{Spikes per image in Conv2. Shown in the inset zoomed to show images
1000 to 1020.}%
\label{l4_spikes_per_img_no_inh}%
\end{figure}

\begin{figure}[!t]%
\centering
\includegraphics[
height=1.8in,
width=3.4in
]%
{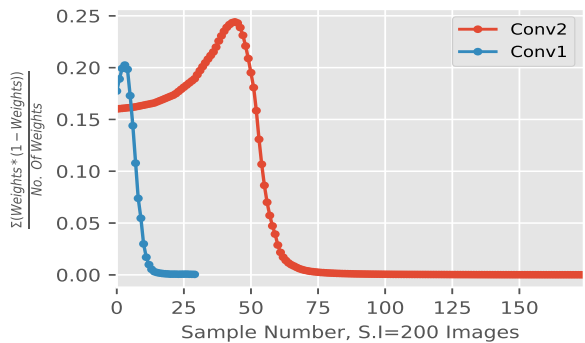}%
\caption{Convergence plots for layers L2 (Conv1) and Conv2.}%
\label{l2l4_weight_convergence_no_inh}%
\end{figure}

\begin{figure}[!t]%
\centering
\includegraphics[
height=1.8in,
width=3.4in
]%
{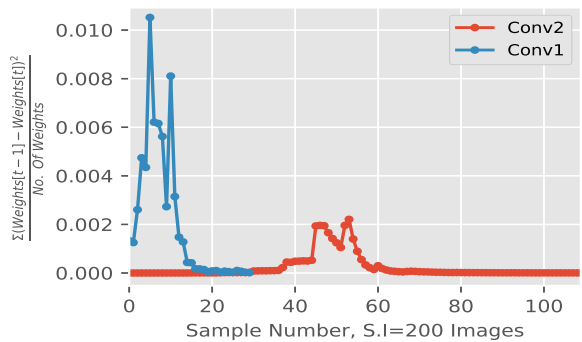}%
\caption{Temporal difference plots for layers L2 (Conv1) and Conv2.}%
\label{l2l4_weight_difference_no_inh}%
\end{figure}

The \texttt{net4.feature\_convergence} method was used to generate
plots shown in  Figures \ref{l2l4_weight_convergence_no_inh}
and \ref{l2l4_weight_difference_no_inh} and these plots show that the synapses of the Conv2 layer  converge slower than layer L2 (Conv1). Such behavior is expected as
the Conv2 layer tries to learn features that are more complex than that
of the features in layer L2 (Conv1). For a Conv2 layer trained with
spikes collected without lateral inhibition in layer L3 (Pool 1) an early
stopping mechanism based on the temporal differences ca be used (see \cite{vaila2019a})
and an example plot is shown in Figure \ref{l2l4_weight_difference_no_inh}).
The \texttt{net4.spike\_statistics()} method was used to generate
plots shown in Figure \ref{l4_bar_chart_weight_updates_no_inh} and Figure
\ref{l4_spikes_per_img_no_inh}.%

\begin{figure}[!t]%
\centering
\includegraphics[
height=2.0in,
width=3.4in
]%
{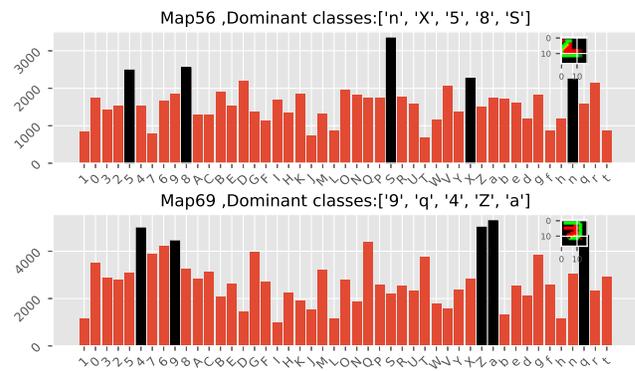}%
\caption{Spikes per map per label in Pool 2. Highlighted (in black) are the
classes that resulted in most number of spikes in a particular feature map.
Feature learned by a map is shown in the inset.}%
\label{l5_spikes_per_map_per_label_no_inh_ieee}%
\end{figure}

The \texttt{net4.animation()} method is used to produce an
animation of Conv2 features\footnote{\url{https://www.youtube.com/watch?v=xtywjRcHmaI}}.
Spikes were collected in Pool 2 layer by fixing the weights of
layer L4 (Conv1). The spikes per map per label in Pool 2 layer are shown in
Figure \ref{l5_spikes_per_map_per_label_no_inh_ieee} similar to layer L3 (Pool 1). The accumulated spikes in Pool 2 can be converted into spike feature vectors using the method
\texttt{network}.\texttt{make\_feature\_vecs()}with the resulting feature
vectors classified using the class\texttt{\ inputlayerclass}.

\begin{figure*}[!t]
\centering
\includegraphics[
height=1.5in,
width=6.5in
]
{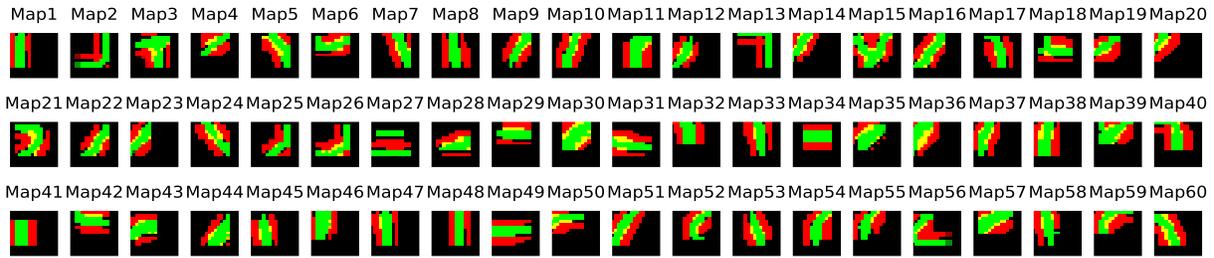}
\caption{Evolved features in L4 (Conv2).}%
\label{l4_features_no_inh_ieee}%
\end{figure*}

The \texttt{net4.feature\_visualization()} method was used to
generate the plots of some of the features of the Conv2 layer as shown in
Figure \ref{l4_features_no_inh_ieee}.%

\subsection{Conv2 and Pool 2 with lateral inhibition in Pool 1}
The synapses in L4 (Conv2) are also trained using the spikes
collected in L3 (Pool 1) with lateral inhibition.

\begin{python}
nTrain_images = 60000, l4_maps = 200 
#conv2 needs more training if lateral inh in\
#pool1 is False
size = net3.pool_spike_tensor.shape[0]
input_channels = net3.pool_spike_tensor.shape[2]
inputs = \
net3.pool_spike_tensor[:,:,:,0:nTrain_images*tsteps]
net5 = network.Network(pool_lateral_inh=False,\
inputs=inputs, A_plus=0.0002, debug=False,\
output_channels=l4_maps, size=size,\
input_channels=input_channels, lr_inc_rate=1500,\
sample_interval=200, train=True,\
threshold=15.0, inh_reg=3, epochs=2)
net5.feedforward()
\end{python}

\begin{figure}%
\centering
\includegraphics[
height=1.7in,
width=3.3in
]%
{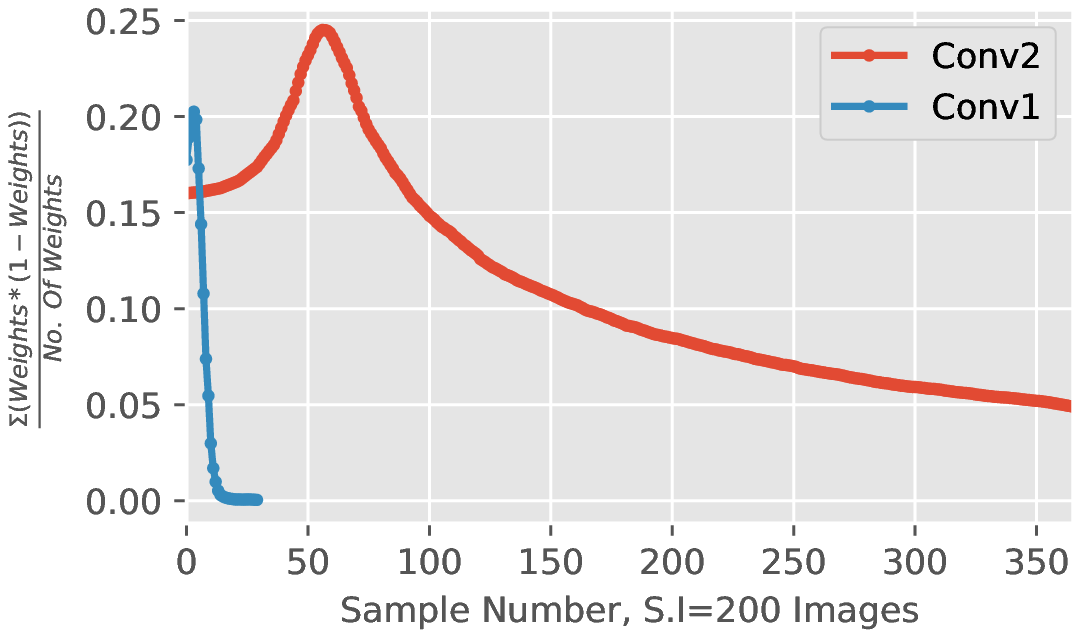}%
\caption{Temporal difference plot for Conv2 synapses when trained with spikes
collected from L3 layer (Pool 1) with lateral inhibition. }%
\label{l2l4_weight_convergence_inh}%
\end{figure}

\begin{figure}%
\centering
\includegraphics[
height=1.7in,
width=3.3in
]%
{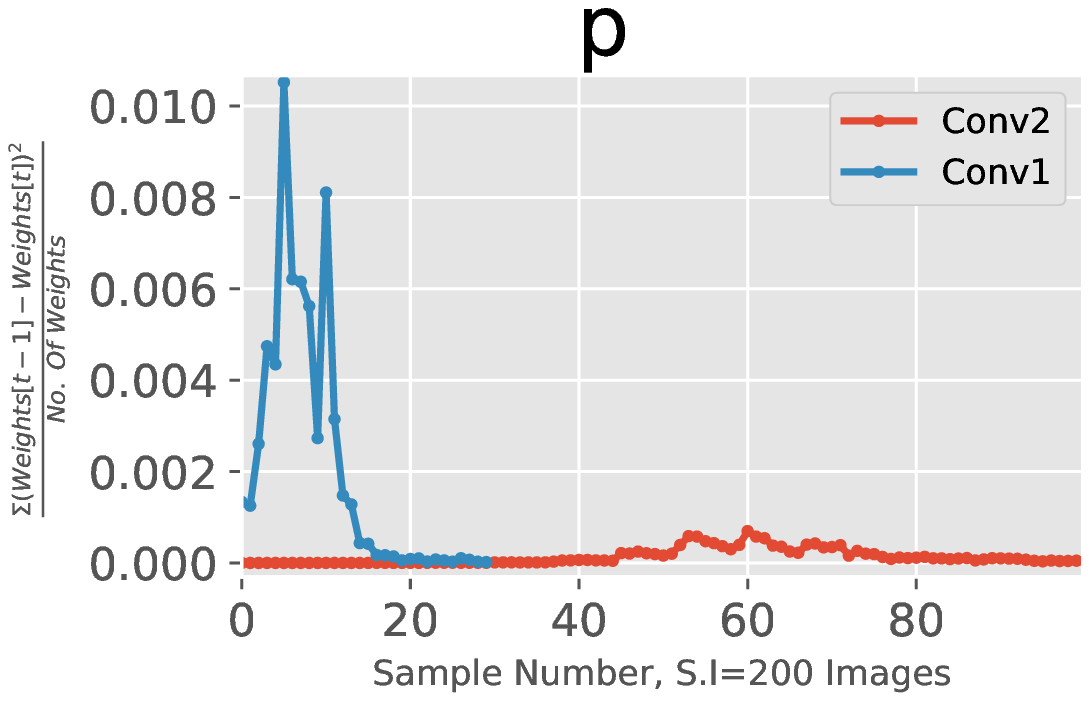}%
\caption{Temporal difference plot for Conv2 synapses when trained with spikes
collected from L3 layer (Pool 1) with lateral inhibition. }%
\label{l2l4_weight_difference_inh}%
\end{figure}

Convergence plots for the weights in the Conv2 layer (with
inhibition in pool 1) are shown in Figure \ref{l2l4_weight_convergence_inh}
and Figure \ref{l2l4_weight_difference_inh}

\begin{figure}
\centering
\includegraphics[
height=1.7in,
width=3.3in
]%
{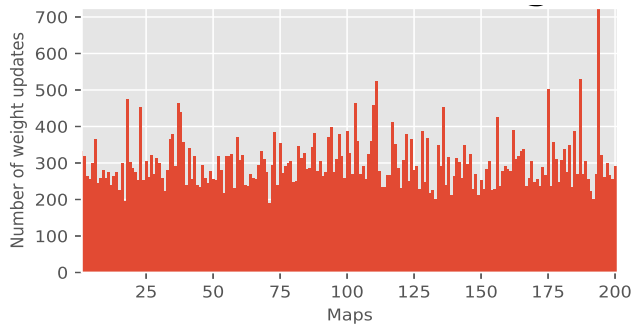}%
\caption{Number of spikes that resulted in STDP weight updates per map.}%
\label{l4_bar_chart_weight_updates_inh}%
\end{figure}

\begin{figure}
\centering
\includegraphics[
height=1.7in,
width=3.3in
]%
{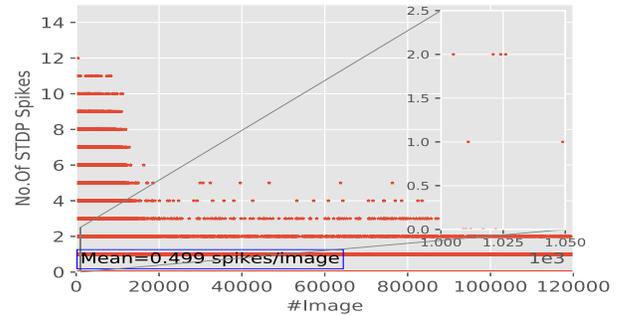}%
\caption{Spikes per image in layer L4. Notice that mean spikes per image is
up to $10\times$ lesser than the case where layer L4 (Conv2) was trained
without lateral inhibition in layer L3 (Pool 1).}%
\label{l4_spikes_per_img_inh}%
\end{figure}

Spike statistics for the Conv2 layer are given in Figure
\ref{l4_bar_chart_weight_updates_inh}\ and Figure \ref{l4_spikes_per_img_inh}.

\begin{figure*}[!t]
\centering
\includegraphics[
height=1.5in,
width=6.5in
]
{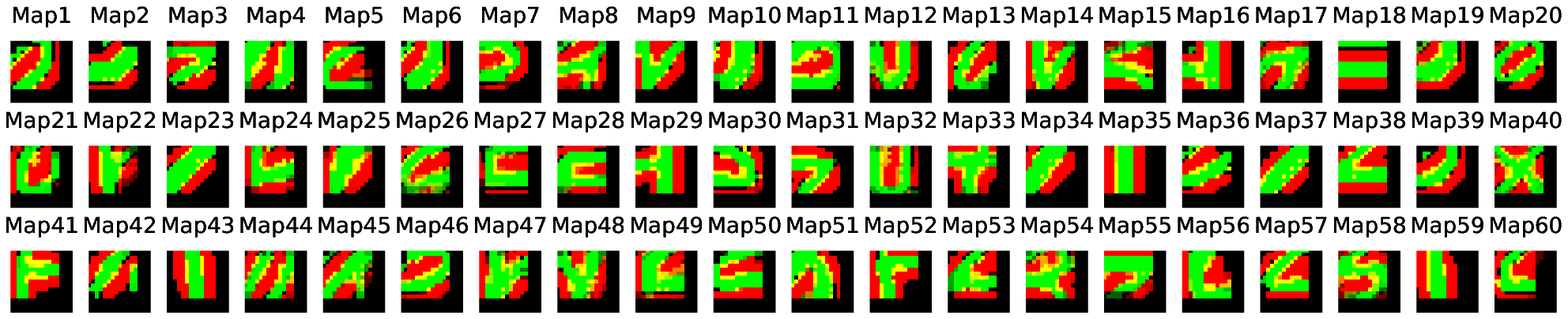}
\caption{Final synapses of L4 layer (Conv2) at the end of training. Notice
that the evolved features are more complex when compared to those in Figure \ref{l4_features_no_inh_ieee}.}%
\label{l4_features_inh_ieee}%
\end{figure*}

The weights (synapses) at the end of the training of Conv2 are given
in Figure \ref{l4_features_inh_ieee}. The animation of the feature evolution for
the Conv2 weights is given at \footnote{\url{https://www.youtube.com/watch?v=rL51343G-Yk&t=24s}}.

\begin{figure}[H]
\centering
\includegraphics[
height=2.0in,
width=3.4in
]%
{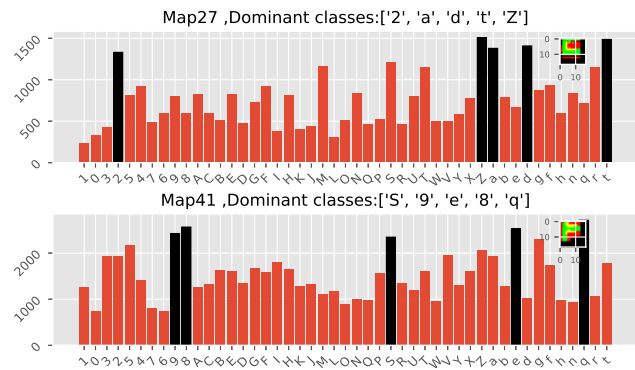}%
\caption{Spikes per map per label in L5. Highlighted (in black) are the
classes that resulted in the most number of spikes in a particular feature map.
Feature learned by a map is shown in the inset.}%
\label{l5_spikes_per_map_per_label_inh}%
\end{figure}

Similar to spikes per map per label in layer L3 (Pool 1), a similar
plot for L5 (Pool 2) is also generated and is shown in Figure
\ref{l5_spikes_per_map_per_label_inh}. As the spikes are already accumulated
in L5 (Pool 2), they are converted into spike feature vectors using the method
\texttt{make\_feature\_vecs} as described for layer L3 (Pool 1).

\section{Conclusions}
We have shown that by combining biologically inspired methods (latency encoded
spikes, STDP) and backpropagation (with surrogate gradients) one can achieve
up to 85.3\% accuracy with the \textsc{EMNIST} dataset. This was done using
backpropagation only in the classification layers of the network as the
classification layers are decoupled from the feature extraction layers. The
accuracy achieved here is quite comparable to the 85.57\% accuracy reported in
\cite{jin} which used rate encoded (Poisson) input spikes in a network with
one hidden layer of 800 neurons and with backpropagation performed in all the
layers. We also demonstrated an accuracy of 94.5\% when the classifier was
given the information that an input image was a letter (upper or lower case)
or a digit. As discussed in the paper, this conditioning was considered
because of the indistinguishability of some of the data between some of the
classes (e.g., \{0,O\} in Figure
\ref{misclass_regular_backprop_softmax_selected_v2_1}).

Using a conventional deep convolution network Shawon et al \cite{shawon}
report an accuracy of 90.59\% on the balanced EMNIST (see also the survey
paper \cite{Baldominos}). The deep network in \cite{shawon} consisted of 6
convolution layers, a hidden layer with 64 neurons followed by a
classification layer.

The computational advantages of using binary activations with respect to a
custom hardware implementation \cite{vishal2018} \cite{wu} \cite{wu2015}
\cite{wu2018} \cite{ivans} \cite{Dahl2018} of bio-inspired neural networks were also presented. The
software tool \textsc{SpykeFlow} was developed and used to simulate the
network and visualize its results. The \textsc{SpykeFlow} tool provides
useful information to the users about feature extraction and spike activity
at various stages of the spiking network serving
as a diagnostic tool of learning in SNNs.

\section{Acknowledgments}
Ruthvik Vaila is thankful to the Elecctrical and Computer Engineering
Department, Boise State University, Idaho, USA for providing him a graduate
assistantship (GA) to carry out this work.

\bibliographystyle{IEEEtran}
\bibliography{ieee_tetci}

% Generated by IEEEtran.bst, version: 1.14 (2015/08/26)
\begin{thebibliography}{10}
\providecommand{\url}[1]{#1}
\csname url@samestyle\endcsname
\providecommand{\newblock}{\relax}
\providecommand{\bibinfo}[2]{#2}
\providecommand{\BIBentrySTDinterwordspacing}{\spaceskip=0pt\relax}
\providecommand{\BIBentryALTinterwordstretchfactor}{4}
\providecommand{\BIBentryALTinterwordspacing}{\spaceskip=\fontdimen2\font plus
\BIBentryALTinterwordstretchfactor\fontdimen3\font minus
  \fontdimen4\font\relax}
\providecommand{\BIBforeignlanguage}[2]{{%
\expandafter\ifx\csname l@#1\endcsname\relax
\typeout{** WARNING: IEEEtran.bst: No hyphenation pattern has been}%
\typeout{** loaded for the language `#1'. Using the pattern for}%
\typeout{** the default language instead.}%
\else
\language=\csname l@#1\endcsname
\fi
#2}}
\providecommand{\BIBdecl}{\relax}
\BIBdecl

\bibitem{delorme2001}
\BIBentryALTinterwordspacing
A.~Delorme, L.~Perrinet, and S.~J. Thorpe, ``{Networks of integrate-and-fire
  neurons using Rank Order Coding B: Spike timing dependent plasticity and
  emergence of orientation selectivity},'' \emph{Neurocomputing}, vol. 38-40,
  pp. 539 -- 545, 2001, computational Neuroscience: Trends in Research 2001.
  [Online]. Available:
  \url{http://www.sciencedirect.com/science/article/pii/S0925231201004039}
\BIBentrySTDinterwordspacing

\bibitem{Kiselev}
M.~{Kiselev}, ``Rate coding vs. temporal coding - is optimum between?'' in
  \emph{2016 International Joint Conference on Neural Networks (IJCNN)}, July
  2016, pp. 1355--1359.

\bibitem{Markram}
\BIBentryALTinterwordspacing
H.~Markram, W.~Gerstner, and P.~J. Sj\"{o}str\"{o}m, ``{Spike-Timing-Dependent
  Plasticity: A Comprehensive Overview},'' \emph{Frontiers in Synaptic
  Neuroscience}, vol.~4, p.~2, 2012. [Online]. Available:
  \url{https://www.frontiersin.org/article/10.3389/fnsyn.2012.00002}
\BIBentrySTDinterwordspacing

\bibitem{Masquelier2008}
\BIBentryALTinterwordspacing
T.~Masquelier, R.~Guyonneau, and S.~J. Thorpe, ``{Spike Timing Dependent
  Plasticity Finds the Start of Repeating Patterns in Continuous Spike
  Trains},'' \emph{PLOS ONE}, vol.~3, no.~1, pp. 1--9, 01 2008. [Online].
  Available: \url{https://doi.org/10.1371/journal.pone.0001377}
\BIBentrySTDinterwordspacing

\bibitem{bottou2016}
L.~Bottou, F.~E. Curtis, and J.~Nocedal, ``{Optimization Methods for
  Large-Scale Machine Learning},'' 2016.

\bibitem{Lecun98}
Y.~{Lecun}, L.~{Bottou}, Y.~{Bengio}, and P.~{Haffner}, ``Gradient-based
  learning applied to document recognition,'' \emph{Proceedings of the IEEE},
  vol.~86, no.~11, pp. 2278--2324, Nov 1998.

\bibitem{frenkel2019}
C.~Frenkel, M.~Lefebvre, and D.~Bol, ``Learning without feedback: Direct random
  target projection as a feedback-alignment algorithm with layerwise
  feedforward training,'' 2019.

\bibitem{Whittington}
\BIBentryALTinterwordspacing
J.~C.~R. Whittington and R.~Bogacz, ``{Theories of Error Back-Propagation in
  the Brain},'' \emph{Trends in Cognitive Sciences}, vol.~23, no.~3, pp.
  235--250, Jan. 2020. [Online]. Available:
  \url{https://doi.org/10.1016/j.tics.2018.12.005}
\BIBentrySTDinterwordspacing

\bibitem{GROSSBERG198723}
\BIBentryALTinterwordspacing
S.~Grossberg, ``Competitive learning: From interactive activation to adaptive
  resonance,'' \emph{Cognitive Science}, vol.~11, no.~1, pp. 23 -- 63, 1987.
  [Online]. Available:
  \url{http://www.sciencedirect.com/science/article/pii/S0364021387800253}
\BIBentrySTDinterwordspacing

\bibitem{Liao}
Q.~{Liao}, J.~Z. {Leibo}, and T.~{Poggio}, ``{How Important is Weight Symmetry
  in Backpropagation?}'' \emph{arXiv e-prints}, p. arXiv:1510.05067, Oct 2015.

\bibitem{Aidan}
\BIBentryALTinterwordspacing
A.~Rocke, \emph{The weight transport problem}.\hskip 1em plus 0.5em minus
  0.4em\relax paulispace, Jun 2017. [Online]. Available:
  \url{https://paulispace.com/deep/learning/2017/06/30/weight-transport.html}
\BIBentrySTDinterwordspacing

\bibitem{Lillicrap}
T.~Lillicrap, D.~Cownden, D.~Tweed, and C.~J.~Akerman, ``Random synaptic
  feedback weights support error backpropagation for deep learning,''
  \emph{Nature Communications}, vol.~7, p. 13276, 11 2016.

\bibitem{Neftci}
\BIBentryALTinterwordspacing
E.~O. Neftci, C.~Augustine, S.~Paul, and G.~Detorakis, ``{Event-Driven Random
  Back-Propagation: Enabling Neuromorphic Deep Learning Machines},''
  \emph{Frontiers in Neuroscience}, vol.~11, p. 324, 2017. [Online]. Available:
  \url{https://www.frontiersin.org/article/10.3389/fnins.2017.00324}
\BIBentrySTDinterwordspacing

\bibitem{mnist}
Y.~LeCun, C.~Cortes, and C.~Burges, ``Mnist handwritten digit database,''
  \emph{ATT Labs [Online]. Available: http://yann. lecun. com/exdb/mnist},
  vol.~2, 2010.

\bibitem{cifar10}
\BIBentryALTinterwordspacing
A.~Krizhevsky, V.~Nair, and G.~Hinton, ``{The CIFAR-10 dataset},'' May 2012.
  [Online]. Available: \url{https://www.cs.toronto.edu/~kriz/cifar.html}
\BIBentrySTDinterwordspacing

\bibitem{courbariaux2016}
M.~Courbariaux, I.~Hubara, D.~Soudry, R.~El-Yaniv, and Y.~Bengio, ``{Binarized
  Neural Networks: Training Deep Neural Networks with Weights and Activations
  Constrained to +1 or -1},'' 2016.

\bibitem{Panda}
C.~Lee, P.~Panda, G.~Srinivasan, and K.~Roy, ``{Training Deep Spiking
  Convolutional Neural Networks With {STDP}-Based Unsupervised Pre-training
  Followed by Supervised Fine-Tuning},'' \emph{Frontiers in Neuroscience},
  vol.~12, p. 435, 08 2018.

\bibitem{imagenet}
J.~Deng, W.~Dong, R.~Socher, L.-J. Li, K.~Li, and L.~Fei-Fei, ``{ImageNet: A
  Large-Scale Hierarchical Image Database},'' 2009.

\bibitem{Anwani}
N.~Anwani and B.~Rajendran, ``Norm{AD} - {N}ormalized {A}pproximate {D}escent
  based supervised learning rule for spiking neurons,'' in \emph{2015
  International Joint Conference on Neural Networks (IJCNN)}, July 2015, pp.
  1--8.

\bibitem{Lee}
\BIBentryALTinterwordspacing
J.~H. Lee, T.~Delbruck, and M.~Pfeiffer, ``{Training Deep Spiking Neural
  Networks Using Backpropagation},'' \emph{Frontiers in Neuroscience}, vol.~10,
  p. 508, 2016. [Online]. Available:
  \url{https://www.frontiersin.org/article/10.3389/fnins.2016.00508}
\BIBentrySTDinterwordspacing

\bibitem{Tavanaei2018}
A.~Tavanaei, Z.~Kirby, and A.~Maida, ``{Training Spiking ConvNets by STDP and
  Gradient Descent},'' \emph{2018 International Joint Conference on Neural
  Networks (IJCNN)}, pp. 1--8, 07 2018.

\bibitem{thiele2019}
J.~C. Thiele, O.~Bichler, A.~Dupret, S.~Solinas, and G.~Indiveri, ``{A Spiking
  Network for Inference of Relations Trained with Neuromorphic
  Backpropagation},'' p. arXiv:1903.04341, Mar 2019.

\bibitem{Lungu}
B.~{Rueckauer}, I.-A. {Lungu}, Y.~{Hu}, and M.~{Pfeiffer}, ``{Theory and Tools
  for the Conversion of Analog to Spiking Convolutional Neural Networks},''
  \emph{arXiv e-prints}, p. arXiv:1612.04052, Dec 2016.

\bibitem{Masquelier2017}
T.~Masquelier, ``Spike-based computing and learning in brains, machines, and
  visual systems in particular (hdr report),'' Ph.D. dissertation,
  Universit{\'e} Toulouse III - Paul Sabatier, 10 2017.

\bibitem{comsa2019}
I.~M. {Comsa}, K.~{Potempa}, L.~{Versari}, T.~{Fischbacher}, A.~{Gesmundo}, and
  J.~{Alakuijala}, ``{Temporal coding in spiking neural networks with alpha
  synaptic function},'' \emph{arXiv e-prints}, p. arXiv:1907.13223, Jul 2019.

\bibitem{Gardner}
\BIBentryALTinterwordspacing
B.~Gardner and A.~Gr{\"{u}}ning, ``{Supervised Learning in Spiking Neural
  Networks for Precise Temporal Encoding},'' \emph{PLOS ONE}, vol.~11, no.~8,
  pp. 1--28, 08 2016. [Online]. Available:
  \url{https://doi.org/10.1371/journal.pone.0161335}
\BIBentrySTDinterwordspacing

\bibitem{kheradpisheh2019s4nn}
S.~R. Kheradpisheh and T.~Masquelier, ``S4nn: temporal backpropagation for
  spiking neural networks with one spike per neuron,'' 2019.

\bibitem{mostafa2018}
H.~{Mostafa}, ``{Supervised Learning Based on Temporal Coding in Spiking Neural
  Networks},'' \emph{IEEE Transactions on Neural Networks and Learning
  Systems}, vol.~29, no.~7, pp. 3227--3235, July 2018.

\bibitem{panda2019}
P.~Panda, A.~Aketi, and K.~Roy, ``{Towards Scalable, Efficient and Accurate
  Deep Spiking Neural Networks with Backward Residual Connections, Stochastic
  Softmax and Hybridization},'' 2019.

\bibitem{Kheradpisheh_2016}
\BIBentryALTinterwordspacing
S.~R. Kheradpisheh, M.~Ganjtabesh, S.~J. Thorpe, and T.~Masquelier,
  ``{STDP}-based spiking deep convolutional neural networks for object
  recognition,'' \emph{Neural Networks}, vol.~99, pp. 56 -- 67, 2018. [Online].
  Available:
  \url{http://www.sciencedirect.com/science/article/pii/S0893608017302903}
\BIBentrySTDinterwordspacing

\bibitem{Kheradpisheh_2016b}
\BIBentryALTinterwordspacing
S.~R. Kheradpisheh, M.~Ganjtabesh, and T.~Masquelier, ``Bio-inspired
  unsupervised learning of visual features leads to robust invariant object
  recognition,'' \emph{Neurocomputing}, vol. 205, pp. 382 -- 392, 2016.
  [Online]. Available:
  \url{http://www.sciencedirect.com/science/article/pii/S0925231216302880}
\BIBentrySTDinterwordspacing

\bibitem{Kheradpisheh}
S.~R. Kheradpisheh, private communication.

\bibitem{edvs}
J.~Conradt, R.~Berner, M.~Cook, and T.~Delbruck, ``{An embedded AER dynamic
  vision sensor for low-latency pole balancing},'' in \emph{2009 IEEE 12th
  International Conference on Computer Vision Workshops, ICCV Workshops}, Sep.
  2009, pp. 780--785.

\bibitem{vaila2019}
R.~{Vaila}, J.~{Chiasson}, and V.~{Saxena}, ``{Deep Convolutional Spiking
  Neural Networks for Image Classification},'' \emph{arXiv e-prints}, p.
  arXiv:1903.12272, Mar 2019.

\bibitem{vaila2019a}
\BIBentryALTinterwordspacing
R.~Vaila, J.~Chiasson, and V.~Saxena, ``{Feature Extraction Using Spiking
  Convolutional Neural Networks},'' in \emph{Proceedings of the International
  Conference on Neuromorphic Systems}, ser. ICONS ’19.\hskip 1em plus 0.5em
  minus 0.4em\relax New York, NY, USA: Association for Computing Machinery,
  2019. [Online]. Available: \url{https://doi.org/10.1145/3354265.3354279}
\BIBentrySTDinterwordspacing

\bibitem{Nielsen}
\BIBentryALTinterwordspacing
M.~A. Nielsen, \emph{Neural {N}etworks and {D}eep {L}earning}.\hskip 1em plus
  0.5em minus 0.4em\relax Determination Press, Jan 2015. [Online]. Available:
  \url{http://neuralnetworksanddeeplearning.com/}
\BIBentrySTDinterwordspacing

\bibitem{emnist}
G.~{Cohen}, S.~{Afshar}, J.~{Tapson}, and A.~{van Schaik}, ``{{EMNIST: an
  extension of MNIST to handwritten letters}},'' \emph{arXiv e-prints}, p.
  arXiv:1702.05373, Feb. 2017.

\bibitem{wu}
X.~{Wu}, V.~{Saxena}, and K.~{Zhu}, ``{Homogeneous Spiking Neuromorphic System
  for Real-World Pattern Recognition},'' \emph{IEEE Journal on Emerging and
  Selected Topics in Circuits and Systems}, vol.~5, no.~2, pp. 254--266, June
  2015.

\bibitem{wu2015}
X.~{Wu}, V.~{Saxena}, K.~{Zhu}, and S.~{Balagopal}, ``{A CMOS Spiking Neuron
  for Brain-Inspired Neural Networks With Resistive Synapses and In Situ
  Learning},'' \emph{IEEE Transactions on Circuits and Systems II: Express
  Briefs}, vol.~62, no.~11, pp. 1088--1092, Nov 2015.

\bibitem{vaila2018}
\BIBentryALTinterwordspacing
V.~S. Ruthvik~Vaila, John~Chiasson, ``Spiking {CNN}s with {PYNN} and
  {NEURON},'' in \emph{NICE Workshop Series}.\hskip 1em plus 0.5em minus
  0.4em\relax Portland, Oregon, USA: Intel, Feb. 2019. [Online]. Available:
  \url{https://www.researchgate.net/publication/335635588_Spiking_CNNs_with_PYNN_and_NEURON}
\BIBentrySTDinterwordspacing

\bibitem{neuron}
\BIBentryALTinterwordspacing
M.~Hines and T.~Carnevale, \emph{{NEURON Simulation Environment}}.\hskip 1em
  plus 0.5em minus 0.4em\relax New York, NY: Springer New York, 2013, pp. 1--8,
  {In Encyclopedia of Computational Neuroscience, Jaeger, Dieter and Jung,
  Ranu, Editors}. [Online]. Available:
  \url{https://doi.org/10.1007/978-1-4614-7320-6_795-1}
\BIBentrySTDinterwordspacing

\bibitem{nengo}
\BIBentryALTinterwordspacing
T.~Bekolay, J.~Bergstra, E.~Hunsberger, T.~DeWolf, T.~Stewart, D.~Rasmussen,
  X.~Choo, A.~Voelker, and C.~Eliasmith, ``{Nengo: a Python tool for building
  large-scale functional brain models},'' \emph{Frontiers in Neuroinformatics},
  vol.~7, p.~48, 2014. [Online]. Available:
  \url{https://www.frontiersin.org/article/10.3389/fninf.2013.00048}
\BIBentrySTDinterwordspacing

\bibitem{spyketorch}
\BIBentryALTinterwordspacing
M.~Mozafari, M.~Ganjtabesh, A.~Nowzari-Dalini, and T.~Masquelier,
  ``{SpykeTorch: Efficient Simulation of Convolutional Spiking Neural Networks
  With at Most One Spike per Neuron},'' \emph{Frontiers in Neuroscience},
  vol.~13, p. 625, 2019. [Online]. Available:
  \url{https://www.frontiersin.org/article/10.3389/fnins.2019.00625}
\BIBentrySTDinterwordspacing

\bibitem{pytorch}
\BIBentryALTinterwordspacing
A.~Paszke, S.~Gross, F.~Massa, A.~Lerer, J.~Bradbury, G.~Chanan, T.~Killeen,
  Z.~Lin, N.~Gimelshein, L.~Antiga, A.~Desmaison, A.~Kopf, E.~Yang, Z.~DeVito,
  M.~Raison, A.~Tejani, S.~Chilamkurthy, B.~Steiner, L.~Fang, J.~Bai, and
  S.~Chintala, ``{PyTorch: An Imperative Style, High-Performance Deep Learning
  Library},'' in \emph{Advances in Neural Information Processing Systems 32},
  H.~Wallach, H.~Larochelle, A.~Beygelzimer, F.~d'Alch\'{e} Buc, E.~Fox, and
  R.~Garnett, Eds.\hskip 1em plus 0.5em minus 0.4em\relax Curran Associates,
  Inc., 2019, pp. 8024--8035. [Online]. Available:
  \url{http://papers.neurips.cc/paper/9015-pytorch-an-imperative-style-high-performance-deep-learning-library.pdf}
\BIBentrySTDinterwordspacing

\bibitem{numpy}
\BIBentryALTinterwordspacing
S.~v.~d. Walt, S.~C. Colbert, and G.~Varoquaux, ``{The NumPy Array: A Structure
  for Efficient Numerical Computation},'' \emph{Computing in Science \&
  Engineering}, vol.~13, no.~2, pp. 22--30, 2011. [Online]. Available:
  \url{https://aip.scitation.org/doi/abs/10.1109/MCSE.2011.37}
\BIBentrySTDinterwordspacing

\bibitem{tensorflow2015}
\BIBentryALTinterwordspacing
M.~Abadi, A.~Agarwal, P.~Barham, E.~Brevdo, Z.~Chen, C.~Citro, G.~S. Corrado,
  A.~Davis, J.~Dean, M.~Devin, S.~Ghemawat, I.~Goodfellow, A.~Harp, G.~Irving,
  M.~Isard, Y.~Jia, R.~Jozefowicz, L.~Kaiser, M.~Kudlur, J.~Levenberg,
  D.~Man\'{e}, R.~Monga, S.~Moore, D.~Murray, C.~Olah, M.~Schuster, J.~Shlens,
  B.~Steiner, I.~Sutskever, K.~Talwar, P.~Tucker, V.~Vanhoucke, V.~Vasudevan,
  F.~Vi\'{e}gas, O.~Vinyals, P.~Warden, M.~Wattenberg, M.~Wicke, Y.~Yu, and
  X.~Zheng, ``{TensorFlow: Large-Scale Machine Learning on Heterogeneous
  Systems},'' 2015, software available from tensorflow.org. [Online].
  Available: \url{https://www.tensorflow.org/}
\BIBentrySTDinterwordspacing

\bibitem{Matplotlib}
J.~D. Hunter, ``{Matplotlib: A 2D graphics environment},'' \emph{Computing in
  Science \& Engineering}, vol.~9, no.~3, pp. 90--95, 2007.

\bibitem{jin}
Y.~Jin, P.~Li, and W.~Zhang, ``{Hybrid Macro/Micro Level Backpropagation for
  Training Deep Spiking Neural Networks},'' \emph{arXiv-eprints}, 05 2018.

\bibitem{shawon}
A.~{Shawon}, M.~{Jamil-Ur Rahman}, F.~{Mahmud}, and M.~M. {Arefin Zaman},
  ``Bangla handwritten digit recognition using deep cnn for large and unbiased
  dataset,'' in \emph{{2018 International Conference on Bangla Speech and
  Language Processing (ICBSLP)}}, Sep. 2018, pp. 1--6.

\bibitem{Baldominos}
\BIBentryALTinterwordspacing
A.~Baldominos, Y.~Saez, and P.~Isasi, ``{A Survey of Handwritten Character
  Recognition with MNIST and EMNIST},'' \emph{Applied Sciences}, vol.~9,
  no.~15, 2019. [Online]. Available:
  \url{https://www.mdpi.com/2076-3417/9/15/3169}
\BIBentrySTDinterwordspacing

\bibitem{vishal2018}
\BIBentryALTinterwordspacing
V.~Saxena, X.~Wu, I.~Srivastava, and K.~Zhu, ``{T}owards {N}euromorphic
  {L}earning {M}achines {U}sing {E}merging {M}emory {D}evices with
  {B}rain-{L}ike {E}nergy {E}fficiency,'' \emph{Journal of Low Power
  Electronics and Applications}, vol.~8, no.~4, 2018. [Online]. Available:
  \url{http://www.mdpi.com/2079-9268/8/4/34}
\BIBentrySTDinterwordspacing

\bibitem{wu2018}
X.~Wu and V.~Saxena, ``{Dendritic-Inspired Processing Enables Bio-Plausible
  STDP in Compound Binary Synapses},'' \emph{IEEE Transactions on
  Nanotechnology}, vol.~PP, 01 2018.

\bibitem{ivans}
R.~C. {Ivans}, S.~G. {Dahl}, and K.~D. {Cantley}, ``{A Model for R(t) Elements
  and R(t)-Based Spike-Timing-Dependent Plasticity With Basic Circuit
  Examples},'' \emph{IEEE Transactions on Neural Networks and Learning
  Systems}, pp. 1--11, 2019.

\bibitem{Dahl2018}
\BIBentryALTinterwordspacing
S.~G. Dahl, R.~Ivans, and K.~D. Cantley, ``{Modeling Memristor Radiation
  Interaction Events and the Effect on Neuromorphic Learning Circuits},'' in
  \emph{Proceedings of the International Conference on Neuromorphic Systems},
  ser. ICONS '18.\hskip 1em plus 0.5em minus 0.4em\relax New York, NY, USA:
  ACM, 2018, pp. 1:1--1:8. [Online]. Available:
  \url{http://doi.acm.org/10.1145/3229884.3229885}
\BIBentrySTDinterwordspacing

\end{thebibliography}
\end{document}